\pdfoutput=1

\documentclass[11pt]{article}

\usepackage[final]{acl}

\usepackage{times}
\usepackage{latexsym}

\usepackage[T1]{fontenc}
\usepackage{pifont}

\usepackage[utf8]{inputenc}

\usepackage{microtype}
\usepackage[normalem]{ulem}
\useunder{\uline}{\ul}{}
\usepackage{inconsolata}
\usepackage{arydshln}
\usepackage{graphicx}
\usepackage{booktabs}
\usepackage{multirow}
\usepackage{xcolor}
\usepackage{algorithm} 
\usepackage{algpseudocode} 
\usepackage{tikz}
\usepackage{ulem}
\usepackage{amsmath}
\usepackage{array}
\usepackage{colortbl}
\usepackage{makecell}
\usepackage{subcaption}
\usepackage{textcomp}
\usepackage{multirow}
\usepackage{algpseudocode}
\usepackage{xcolor}
\usepackage{setspace}
\usepackage{caption}
\usepackage{setspace}
\usepackage{subcaption}
\usepackage{stfloats}
\definecolor{verylightgray}{rgb}{0.92, 0.92, 0.92}
\definecolor{mycolor}{RGB}{120,180,80}
\definecolor{darkgreen}{RGB}{0,180,0}
\newcommand{\xmark}{\textcolor{red}{\scalebox{1.4}{\ding{56}}}}
\newcommand{\cmark}{\textcolor{darkgreen}{\scalebox{1.4}{\ding{52}}}}
\title{Exploring In-context Example Generation \\ for Machine Translation}

\author{
 \textbf{Dohyun Lee\textsuperscript{1}} \hspace{0.5mm}
 \textbf{Seungil Chad Lee\textsuperscript{1}} \hspace{0.5mm}
 \textbf{Chanwoo Yang\textsuperscript{2}} \hspace{0.5mm}
 \textbf{Yujin Baek\textsuperscript{1}} \hspace{0.5mm}
 \textbf{Jaegul Choo\textsuperscript{1}}
\\
 \textsuperscript{1}KAIST AI,
 \textsuperscript{2}Jeonbuk National University,
\\
\texttt{\{aiclaudev, seungil.lee, yujinbaek, jchoo\}@kaist.ac.kr}
}

\begin{document}
\maketitle
\begin{abstract}
Large language models (LLMs) have demonstrated strong performance across various tasks, leveraging their exceptional in-context learning ability with only a few examples.
Accordingly, the selection of optimal in-context examples has been actively studied in the field of machine translation.
However, these studies presuppose the presence of a demonstration pool with human-annotated pairs, making them less applicable to low-resource languages where such an assumption is challenging to meet.
To overcome this limitation, this paper explores the research direction of in-context example generation for machine translation.
Specifically, we propose Demonstration Augmentation for Translation (DAT), a simple yet effective approach that generates example pairs without relying on any external resources.
This method builds upon two prior criteria, \textit{relevance} and \textit{diversity}, which have been highlighted in previous work as key factors for in-context example selection.
Through experiments and analysis on low-resource languages where human-annotated pairs are scarce, we show that DAT achieves superior translation quality compared to the baselines.
Furthermore, we investigate the potential of progressively accumulating generated pairs during test time to build and reuse a demonstration pool. 
Our implementation is publicly available at \url{https://github.com/aiclaudev/DAT}.
\end{abstract}

\section{Introduction}
\label{sec:introduction}

The recent emergence of large language models (LLMs)~\citep{llama, llama2, gpt4} and in-context learning (ICL)~\citep{icl} has shifted the traditional paradigm of building task-specific models trained on large amounts of human-annotated data, which is costly to collect.
The strength of ICL lies in its versatility, where they achieve outstanding performance across various tasks with just a few task-specific demonstrations~\citep{yao2022react, wei2023chainofthoughtpromptingelicitsreasoning}.
This reduces the training cost on large datasets and allows rapid adaptation to new domains or problems without extensive fine-tuning.
Solving various tasks with a single LLM provides immense value to users seeking assistance in diverse contexts.

\begin{figure}[t]
\centering
  \includegraphics[width=1.0\columnwidth]{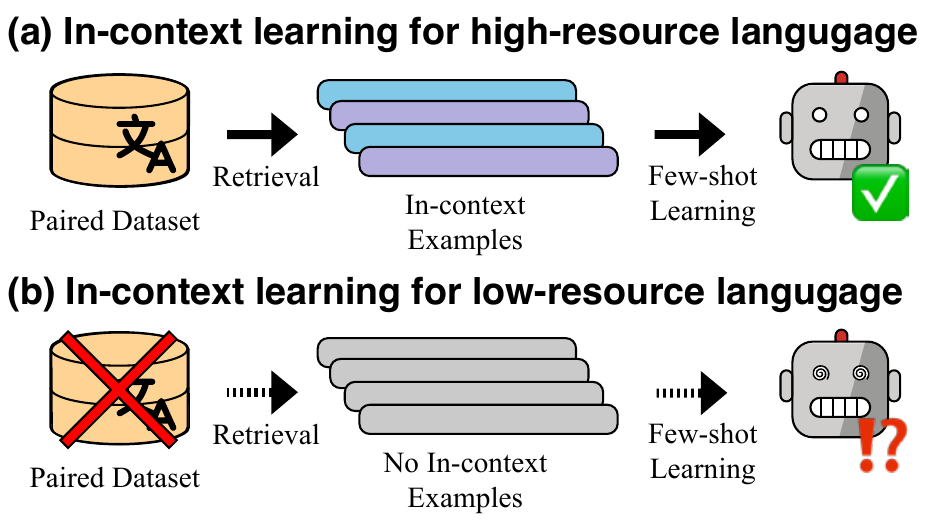}
  \caption{\textbf{Motivation.}
  (a) Previous works on LLM-based in-context learning for translation have primarily focused on selecting in-context examples from paired dataset in high-resource languages.
  (b) However, in the absence of a paired dataset for low-resource languages, how can in-context learning be applied?}
  \label{fig:intro}
\end{figure}

LLMs have also played an increasingly prominent role in the field of machine translation (MT) due to their exceptional linguistic and reasoning capabilities~\citep{domain-specific-moslem-2022, domain-terminology-moslem-2023, prompting-palm-vilar-2023, is-chatgpt-jiao-2023, contextual-refinement-koneru-2024, a-paradigm-xu-2024}.
Notably, ICL has demonstrated high multilingual translation quality using only a few source-target pairs, driving advancements in retrieving optimal examples.
\citet{rbm25} proposed R-BM25, which initially selects the top bm25 candidates and reranks them using an n-gram recall strategy.
\citet{ctqscorer} introduced a neural network trained to select pairs based on the multiple features, including semantic similarity and sentence length.
These works have demonstrated outstanding performance in high-resource languages, retrieving pairs based on the scores between the given user query and source sentences in the demonstration pool, as shown in Figure~\ref{fig:intro} (a).

However, these works on in-context example selection for MT may face challenges in low-resource languages. 
This stems from the fact that previous approaches rely on a critical assumption—namely, the availability of a large pool of human-annotated pairs—which may not hold for low-resource languages. 
For low-resource languages, obtaining a paired corpus for use as demonstrations is challenging due to the limited availability of public datasets and human annotators.
This, in turn, poses a barrier that prevents low-resource languages from fully benefiting from the use of in-context examples, as illustrated in Figure~\ref{fig:intro} (b).
Recently, \citet{self-mining} investigated leveraging LLMs to generate synthetic parallel data as a way to address this obstacle.
However, this approach requires access to the vocabularies of both the source and target languages, as well as unlabeled sentences in the target language.

In this paper, we explore a research direction that aims to enable in-context learning for MT without the use of any external resources, instead drawing solely on the capabilities of the LLM itself.
In pursuit of this goal, we introduce a simple yet effective method, Demonstration Augmentation for Translation (DAT), which utilizes the generative and linguistic capabilities of LLMs.
This approach builds upon the intuitive prior criteria of \textit{relevance} and \textit{diversity}, which are inspired by previous works analyzing desirable in-context examples for MT~\citep{cheng-etal-2022-neural, ontheflymt, bouthors-etal-2024-retrieving}.
To ensure these two criteria, we also utilize maximal marginal relevance~\citep{maximal_marginal_relevance}.

Our experiment focuses on translating from English into low-resource languages—specifically Nepali, Khmer, Pashto, Zulu, and Swahili—for which the lack of extensive annotated datasets presents a realistic constraint.
The results demonstrate the practicability of our easily applicable method in generating pairs that serve as in-context examples, providing valuable clues for user query translation.
One more noteworthy point is that we observe a counterintuitive case where utilizing high-quality fixed pairs results in a severe performance degradation compared to the zero-shot approach.
We investigate this phenomenon with a focus on the relevance between the source side of the pairs and the user queries.
Lastly, we explore an extended method that incrementally accumulates the generated pairs and repurposes them through retrieval method such as R-BM25.

In summary, our contributions are as follows:
\begin{itemize}
    \item To the best of our knowledge, this is the first work to explore in-context example generation specialized for MT without relying on any external resources, such as vocabularies or monolingual corpora.
    \vspace{-2mm}
    \item Experimentes show that DAT boosts the translation quality compared to other baselines, demonstrating its practicality for low-resource languages with scarce human-labeled pairs.
    \vspace{-2mm}
    \item Additional experiments demonstrate that high-quality fixed pairs in low-resource languages can act as noise and highlight DAT's potential for demonstration pool construction.
\end{itemize}

\section{Related Work}
\label{sec:related_work}

\subsection{In-context Learning}
\label{sec:In-context Learning}
In-context learning (ICL) paradigm, originally proposed by ~\citet{icl}, enables LLMs~\citep{llama, llama2, gpt4, llama3.1} to learn new tasks without any parameter updates by providing task-relevant input-output exemplars known as demonstrations~\citep{icexample}.
This paradigm facilitates incorporating human knowledge through task-specific examples into LLMs. It is often more effective than fine-tuning, allowing models to adapt to new cases with reduced data requirements~\citep{iclvsft}.
Previous works have introduced various strategies for constructing ICL prompts, highlighting that adjusting how demonstrations are composed can lead to more efficient solutions across various tasks~\citep{calibrate, retrieve-prompts, structured-prompting, prompting-reliable}.
Moreover, recent studies analyzing the factors influencing ICL performance have further supported its effectiveness~\citep{rethinking, ptcorpora, distproperty, icexample}.
Leveraging ICL, the ability of a single LLM to solve diverse tasks offers significant value in real-world applications.

\subsection{Machine Translation using LLMs}
\label{sec:Machine Translation using LLMs}

Neural machine translation (NMT) models~\citep{nllb2022} are trained with large amounts of high-quality parallel data, which is resource-intensive and costly. 
Consequently, numerous studies have been conducted on leveraging LLMs for machine translation, motivated by the data efficiency benefits offered by ICL.
Extensive research has shown that leveraging zero-shot and few-shot learning techniques achieves translation abilities that match or exceed the performance of traditional NMT models with just a minimal number of demonstrations~\citep{xglm, palm, prompting-palm-vilar-2023, promptingllm, literal, is-chatgpt-jiao-2023}.
Recognizing the importance of effective demonstrations, further studies have focused on optimizing in-context example selection to improve the translation performance of LLMs through ICL, resulting in notable advancements in selection techniques~\citep{rbm25, ctqscorer, submodular, zebaze-etal-2025-context}.
However, these works rely on a large pool of human-annotated demonstrations, which can be impractical in real-world scenarios, especially for translation involving low-resource languages.
To address this limitation, recently, \citet{self-mining} investigated leveraging LLMs to build a synthetic demonstration pool, but the approach requires access to lexical resources for both source and target languages, along with unlabeled data in the target language.
In our method, we leverage only the linguistic capabilities of LLMs to generate in-context examples that enhance machine translation performance, marking the first attempt in the field.

\subsection{Demonstration Augmentation}
\label{sec:Demonstration Augmentation}

Demonstrations play a crucial role in ICL, as they significantly impact model performance by providing task-relevant examples that aid in solving new cases~\citep{active, overcoming, icexample, bouthors-etal-2024-retrieving}.
Research has moved beyond selecting high-quality examples, with growing interest in methods allowing LLMs to generate informative demonstrations autonomously~\citep{sg-icl, z-icl, self-icl, demonstration_augmentation}.
~\citet{human_gen_ICL} confirmed that LLMs can achieve performance comparable to human-curated demonstrations by employing a self-reflective prompting strategy, illustrating that models can independently create examples that inform decision-making without the need for external, human-generated input.
Our work builds on these advancements by exploring how leveraging the reasoning abilities of LLMs to generate source sentences and their translations produced by LLMs enhances translation performance.
Especially, rather than using general approaches, we focus heavily on designing a more specific strategy for MT, eliminating reliance on human intervention or external data.

\section{Method}
\label{sec:method}
\begin{figure*}
    \centering
    \includegraphics[width=\textwidth]{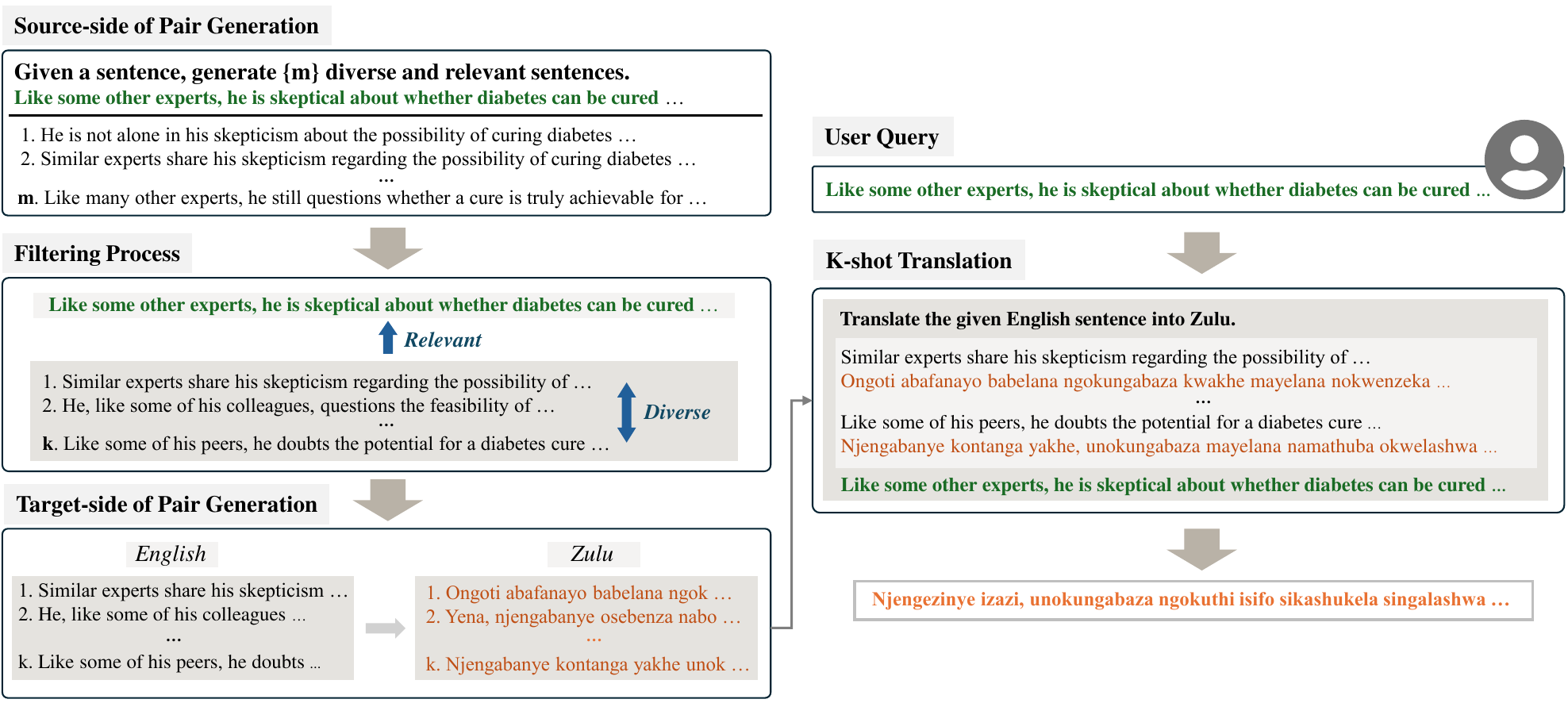}
    \caption{\textbf{An overview of our proposed method}. 
    (1) Upon receiving a translation request for the user's query, the LLM generates \textit{m} source-side sentences that satisfy both relevance and diversity constraints.
(2) A relevant sentence that minimizes redundancy with previously chosen source sentences is iteratively selected and appended to the candidate pool. This process is iterated \textit{k} times.
(3) The LLM then translates each selected sentence, forming source-target pairs.
(4) The final translation is produced through a few-shot learning framework, utilizing the generated pairs as in-context exemplars.
A detailed explanation of our method is provided in Section~\ref{sec:method}.}
    \label{fig:method}
\end{figure*}
\paragraph{Overview.}
We aims to generate pairs for in-context learning in the absence of human-annotated pairs.
At test time, when a user provides a query $q$, LLM generates source-target pairs tailored to $q$ and uses them as in-context examples.
The overall flow of this method is shown in Figure~\ref{fig:method}.
\subsection{Source-side of Pair Generation}
To create source sentences $\mathit{X}=\{x_1, x_2, \cdots, x_m\}$ that provide valuable cues for translating $q$, we draw on previous research that explores the optimal in-context examples for MT~\citep{cheng-etal-2022-neural, ontheflymt}. 
These studies generally propose the following two priors:
\begin{itemize}
    \item \textbf{\textit{Relevance}} refers to the similarity between $q$ and $x_i \in \mathit{X}$, which can include metrics such as n-gram overlap, edit distance, embedding similarity, and bm25 score.
    \item \textbf{\textit{Diversity}} means the distinction between the retrieved examples, based on the intuition that if they are too similar to each other, they will provide redundant clues when translating $q$.
\end{itemize}
Generating $\mathit{X}$ can be easily achieved by zero-shot prompting the LLM (source generator) with $q$ and an instruction that incorporates the two priors. 
The simple instruction is illustrated in Figure~\ref{fig:method} and the detailed prompt used for generating source sentences is presented in Figure~\ref{fig:source_generation_prompt}.

\subsection{Filtering using MMR}
To ensure that the generated sentences satisfy the two prior conditions, we apply filtering and use only $k$ examples.
\textit{Relevance} can be considered by a recall-based n-gram score $\mathrm{R}_n$ between $q$ and $x_i$:
\begin{gather}
    \mathrm{R}_n(q,x_i)=\frac{\sum_{\mathrm{ng} \in f_n(q) \cap f_n(x_i)} Count(\cdot)}{\sum_{\mathrm{ng} \in f_n(q)} Count(\cdot)} \\
    \mathrm{\alpha}(q,x_i)=\frac{1}{4}\sum_{n=1}^{4}\mathrm{R}_n(q,x_i),  
\end{gather}
where $\mathrm{ng}$ means n-gram and $f_n(\cdot)$ refers to the functions that convert a sentence into n-grams.
Moreover, to promote diversity, we select examples using following equation inspired by Maximal Marginal Relevance (MMR)~\citep{maximal_marginal_relevance} and \citet{cheng-etal-2022-neural}.
\begin{gather}
   \underset{x_i \in X \setminus X^*}{\mathrm{argmax}}[\mathrm{\alpha}(q, x_i)-\frac{\lambda}{\lvert X^* \rvert}\sum_{x_j \in X^*}\mathrm{\alpha}(x_j, x_i)], 
\end{gather}
where $X^*$ is a set of already selected sentences and $\lambda$ is a hyperparameter.
MMR filtering process is detailed in Algorithm~\ref{alg:filtering}.
\begin{algorithm}[h]
    \footnotesize  
    \setstretch{1.2} 
    \caption{{Filtering using MMR}}  
    \label{alg:filtering}
    \begin{algorithmic}[1] 
    \State \colorbox{verylightgray}{\textbf{Input}: $q$, $\mathit{X}\!\!=\!\{x_i\}_{i=1}^m$, $k$ (<$m$), $\lambda$}
    \State \colorbox{verylightgray}{\textbf{Output}: Selected Sources $\mathit{X}^*\!\!=\!\{x_i^*\}_{i=1}^k$}
    \noindent\hspace*{-4mm}\rule{1.055\linewidth}{0.4pt}
    \Procedure{Filtering}{$q$, $\mathit{X}$, $k$, $\lambda$}
    \State $\mathit{X}^* \gets \emptyset$
    \While {$\lvert \mathit{X}^* \rvert < k$}
        \For {$x \in \mathit{X} \setminus \mathit{X}^*$}
            \State \textit{Relevance} $\!\gets\!$ $\alpha(q, x)$
            \State \textit{Diversity} $\!\gets\!$ $\frac{-1}{\lvert X^* \rvert}\sum_{x_j \in X^*} \alpha(x_j, x)$
        \EndFor
        \State $x^* \gets \underset{x \in \mathit{X} \setminus \mathit{X}^*}{\mathrm{argmax}} \,(\textit{Relevance} + \lambda \textit{Diversity})$
        \State $\mathit{X}^* \gets X^* \cup \{x^*\}$
    \EndWhile
    \State \Return $\mathit{X}^*$
    \EndProcedure
\end{algorithmic}
\end{algorithm}

\subsection{Target-side of Pair Generation}
After filtering, we need to generate translations for each $x^* \in \mathit{X}^*$.
We have two options for translating the $k$ source sentences: either using the LLM with zero-shot prompting or relying on the NMT model.
LLM has acquired general knowledge across various domains through training on a vast pretraining dataset~\citep{contextual-refinement-koneru-2024} and excels at preserving semantic information~\citep{how-good-hendy-2023}.
For simplicity, we utilize LLM as our target generator.

\subsection{Query Translation}
By generating automatically without relying on any human-curated data, we now obtain $k$ source-target pairs: $D^*=\{(x^*_i, \textsc{LLM}(x^*_i)\}_{i=1}^k$. 
Since these are tailored to the user query, they provide sufficient clues when translating the query. 
The query translator, an LLM, utilizes these demonstrations to perform in-context learning.
\begin{gather}
    \hat{y} = \textsc{LLM}(I, D^*, q),
\end{gather}
where $\hat{y}$ is translated from $q$ and $I$ refers to the instruction (e.g., \texttt{Translate a given <source language> sentence to <target language> sentence}).
Figure~\ref{fig:target_generation_prompt} shows the detailed prompt.

\section{Experimental Setup}
\subsection{Datasets and Languages}
\label{setup: dataset}
We benchmark the Flores dataset~\citep{flores}, focusing on English and five low-resource languages: Nepali, Khmer, Pashto, Zulu, and Swahili. 
To evaluate performance, we conduct experiments on the devtest split, assessing the effectiveness of our approach in these language settings.

\subsection{Evaluation Metrics}
We utilize COMET\footnote{\texttt{Unbabel/wmt22-COMET-da}}~\cite{comet22}, one of the most commonly used evaluation metrics.
We also leverage reference-free COMET\footnote{\texttt{Unbabel/wmt22-cometkiwi-da}}~\cite{rei-etal-2022-cometkiwi} to evaluate the quality of the in-context example in Table~\ref{table:qual_table}.
For a more rigorous evaluation, we use afriCOMET\footnote{\texttt{masakhane/africomet-stl}} and reference-free afriCOMET\footnote{\texttt{masakhane/africomet-qe-stl}}~\citep{africomet} for Zulu and Swahili, as afriCOMET is specialized for African languages.
These metrics are designed to predict human judgments of translation quality.

\subsection{Prompting Setup}
Our proposed method is performed solely through zero-shot prompting, without relying on any pairs or examples.
In our experiments, the number of in-context examples in few-shot prompting is fixed at 4.
A source generator in our method initially generates 10 ($m$) sentences based on the user query with zero-shot prompting. 
Then filtering process remains 4 ($k$) sentences while considering relevance and diversity.  
The prompt template used in the whole experiments is provided in Figure~\ref{fig:source_generation_prompt} and \ref{fig:target_generation_prompt}.

\section{Results and Analyses}
\subsection{Results on Low Resource Languages} \label{sec:5_low_resource}
\begin{table*}[h]
\centering
\small
\begin{tabular}{@{}cccccccc@{}}
\toprule
\textbf{Fixed Pairs} & \textbf{Model} & \textbf{Method} & \textbf{Nepali} & \textbf{Khmer} & \textbf{Pashto} & \textbf{Zulu} & \textbf{Swahili} \\ \midrule
\multirow{4}{*}{\xmark}  & \multirow{2}{*}{\textit{Llama-3.1-8B}}  & Zero-shot       & 72.1            & 62.0           & 53.9            & \textbf{23.3}\rlap{\textsuperscript{*}} & 60.6             \\
                    &                                         & DAT            & \textbf{74.9}\rlap{\textsuperscript{*}}   & \textbf{64.4}\rlap{\textsuperscript{*}}  & \textbf{54.6}   & 22.3          & \textbf{61.8}\rlap{\textsuperscript{*}}    \\ \cmidrule(l){2-8} 
                    & \multirow{2}{*}{\textit{Llama-3.1-70B}} & Zero-shot       & 79.8            & \textbf{72.7}  & 67.5            & 37.8          & 72.9             \\
                    &                                         & DAT            & \textbf{81.1}\rlap{\textsuperscript{*}}   & 72.4           & \textbf{68.3}\rlap{\textsuperscript{*}}   & \textbf{38.3} & \textbf{73.4}\rlap{\textsuperscript{*}}    \\ \midrule \midrule
\multirow{4}{*}{\cmark}  & \multirow{2}{*}{\textit{Llama-3.1-8B}}  & Few-shot        & 75.9            & 65.0           & \textbf{57.9}   & \textbf{24.7}\rlap{\textsuperscript{*}} & 61.3             \\
                    &                                         & DAT            & \textbf{76.4}   & \textbf{66.0}  & 57.3            & 23.3          & \textbf{62.3}\rlap{\textsuperscript{*}}    \\ \cmidrule(l){2-8} 
                    & \multirow{2}{*}{\textit{Llama-3.1-70B}} & Few-shot        & 80.6            & 51.1           & 65.7            & 38.9          & 71.6             \\
                    &                                         & DAT            & \textbf{81.5}\rlap{\textsuperscript{*}}   & \textbf{52.9}\rlap{\textsuperscript{*}}  & \textbf{68.5}\rlap{\textsuperscript{*}}   & \textbf{39.2} & \textbf{72.7}\rlap{\textsuperscript{*}}    \\ \bottomrule
\end{tabular}
\caption{
The experimental results present COMET scores for translating English into five low-resource languages. 
Performance that surpasses the compared method is \textbf{bolded} for clarity, and * indicates statistically significant improvement at p=0.05, using the \texttt{compare-mt} library \citep{compare-mt}.
The "Fixed Pair" column specifies whether a fixed set of human-annotated pairs is utilized during the translation process. 
For a more detailed explanation of this setting, please refer to the experimental configuration in Section~\ref{sec:5_low_resource}.}
\label{table:main_table}
\end{table*}

\paragraph{Experimental Configuration} \label{par:fixed_column}
Table~\ref{table:main_table} reports the COMET scores for translations from English into five low-resource languages. 
In real-world scenarios, these languages typically lack human-curated parallel corpora, which limits the feasibility of approaches such as in-context example selection and few-shot learning explored in previous work. 
As a workaround, one approach involves manually annotating a limited set of translation pairs and integrating them as fixed references during translation, which can serve as anchors and potentially enhance translation quality.
This approach can also be applied to DAT, where these pairs are used during the target sentence generation process rather than when translating the test data.

\paragraph{No Fixed Pairs Setting}
Since this setting does not rely on human-annotated pairs, we investigate whether in-context examples, generated purely from the LLM's intrinsic capabilities, can serve as a catalyst for boosting translation quality beyond a zero-shot baseline.
Our findings indicate that DAT has improved translation quality in most low-resource languages compared to zero-shot translation.
Notably, for the Nepali, Llama-3.1-8B and 70B achieved performance gains of 2.8 and 1.3 points, respectively.
This result shows that LLMs, relying solely on their inherent abilities without external information, enhance translation quality by using self-generated in-context examples.
Consequently, it underscores their potential for generating low-resource language pairs.
\paragraph{Fixed Pairs Setting}
A fixed pair set, while meticulously curated by humans to ensure high quality, may not always align with user queries requiring translation. 
Nevertheless, it can still facilitate performance improvement by offering linguistic cues related to language-specific grammar, syntactic structures, and other idiosyncratic features. 
This is exemplified by Llama-3.1-8B, which achieved a 3.8-point higher COMET score in few-shot method for the Nepali compared to the zero-shot method.
Furthermore, DAT, which integrates a fixed human-curated data into the target sentence translation process, exhibits superior translation adequacy compared to the few-shot method across the majority of languages.
This suggests that DAT enhances translation performance by dynamically generating semantically and syntactically aligned sentences while leveraging human-validated data to build high-fidelity translation pairs, resulting in greater adequacy and fluency.
\begin{table*}[h]
\centering
\small
\resizebox{\textwidth}{!}{
\begin{tabular}{@{}ccccccccccccc@{}}
\toprule
\multicolumn{1}{c|}{}                                  & \multicolumn{4}{c|}{\textbf{Nepali}}                                              & \multicolumn{4}{c|}{\textbf{Khmer}}                                               & \multicolumn{4}{c}{\textbf{Swahili}}                         \\
\multicolumn{1}{c|}{\multirow{-2}{*}{\textbf{Method}}} & Relev.$\uparrow$     & Uni.$\downarrow$   & Qual.$\uparrow$       & \multicolumn{1}{c|}{COMET$\uparrow$}         & Relev.$\uparrow$     & Uni.$\downarrow$   & Qual.$\uparrow$       & \multicolumn{1}{c|}{COMET$\uparrow$}         & Relev.$\uparrow$     & Uni.$\downarrow$   & Qual.$\uparrow$       & COMET$\uparrow$         \\ \midrule
\multicolumn{13}{c}{\cellcolor[HTML]{EFEFEF}\textit{Llama-3.1-70B}}                                                                                                                                                                                                                           \\ \midrule
\multicolumn{1}{c|}{Retrieval (src)}                         & 7.5           & 5.3          & 80.7          & \multicolumn{1}{c|}{80.5}          & 7.5           & 5.3          & 65.1          & \multicolumn{1}{c|}{61.4}          & 7.5           & 5.3          & 66.5          & 72.2          \\
\multicolumn{1}{c|}{Fixed set (pair)}                             & 3.9           & \textbf{2.8} & \textbf{89.4} & \multicolumn{1}{c|}{80.6}          & 3.9           & \textbf{2.8} & \textbf{85.6} & \multicolumn{1}{c|}{51.1}          & 3.9           & \textbf{2.8} & \textbf{77.2} & 71.6          \\ \midrule
\multicolumn{1}{c|}{DAT}                              & \textbf{25.9} & 24.1         & 82.5          & \multicolumn{1}{c|}{\textbf{81.1}} & \textbf{25.9} & 24.1         & 65.3          & \multicolumn{1}{c|}{\textbf{72.4}} & \textbf{25.9} & 24.1         & 68.9          & \textbf{73.4} \\ \bottomrule
\end{tabular}
}
\caption{
This experiment presents the results of translating English into other low resource languages.
Relevance (Relev.) measures the average n-gram overlap score between the user's query and the source side of an in-context example, while Uniformity (Uni.) evaluates the same averaged score among the source sides of different in-context examples.
Quality (Qual.) is measured using reference-free COMET to evaluate the quality of a single pair. As a final point, COMET represents the score achieved when translating the user query with the given pairs. The best score in each column is highlighted in \textbf{bold}.}
\label{table:qual_table}
\end{table*}
\paragraph{The Backfire of Fixed Human Pair} 
\begin{figure}[t]
\centering
  \includegraphics[width=1.0\columnwidth]{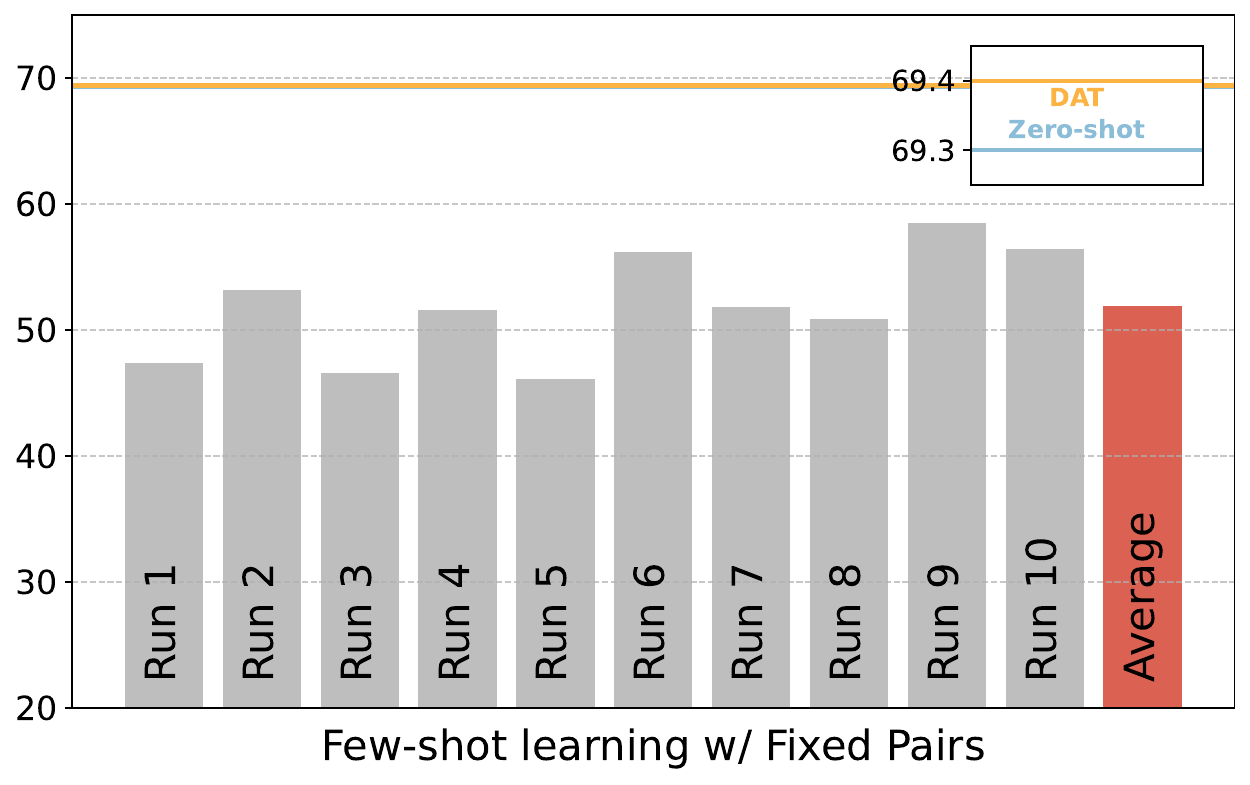}
  \caption{This figure illustrates English-to-Khmer translation results using fixed, high-quality human-aligned pairs. Each run is tested on 100 unique samples, with distinct fixed pair sets used across runs.}
  \label{fig:backfire}
\end{figure}

\begin{table}[h]
\centering
\small
\begin{tabular}{@{}c|cc@{}}
\toprule
\textbf{Method} & \textbf{Off-target Rate} & \textbf{\# of Output Tokens} \\ \midrule
Zero-shot       & 0.0                      & 241.7                        \\
Fixed Pairs           & 0.0                      & 439.3                        \\ \midrule
DAT             & 0.0                      & 231.8                        \\ \bottomrule
\end{tabular}
\caption{These results present the off-target rate, indicating whether the output was translated into the correct language when translating from English to Khmer, along with the number of tokens in the generated sentences. We used \texttt{Google Translate} to identify the language.}
\label{table:off_target}
\end{table}
In this experiment, we observed a counterintuitive result: when translating from English to Khmer using Llama-3.1-70B, the few-shot approach led to a significant drop in translation quality—specifically, a 21.6-point decrease in COMET score compared to the zero-shot baseline.
To assess the robustness of this finding, we ran 10 experiments using various fixed example pairs (Figure~\ref{fig:backfire}) and consistently found that few-shot translation underperformed relative to the zero-shot setting.
As shown in Table~\ref{table:off_target}, our analysis indicates that while the translations remained in the correct target language, incorporating fixed examples often resulted in abnormally long outputs. In some cases, the model repeatedly generated the same strings across different data points.
However, our proposed method avoids these issues, which in turn leads to better translation performance. 
In Table~\ref{table:main_table}, DAT without fixed pairs outperforms the few-shot approach in translation quality for all languages except Zulu.
This raises a research question about whether it is better to use high-quality but fixed pairs that lack relevance to user queries, or moderate-quality pairs that more closely align to them.
A thorough analysis of the underlying reasons is left for future work.

\subsection{Quality of In-context Example}
\label{sec:quality_of_data}
\paragraph{Experimental Configuration}
Previous studies have argued that in-context examples that are both similar to the user query and diverse from one another improve translation performance.
In Table~\ref{table:qual_table}, we explore how the relevance of in-context examples to the user query, the uniformity among them—negatively correlated with diversity—and the intrinsic quality of example pairs affect translation performance.
To investigate this, we consider the following methods: Retrieval, Fixed set, and DAT.
Retrieval selects source sentences from a monolingual pool using R-BM25 scores~\citep{rbm25} and constructs pairs via an LLM, leveraging the accessibility of monolingual data. 
Fixed set relies on a predefined set of human-curated pairs, discussed in Table~\ref{table:main_table}, as in-context examples.
Meanwhile, DAT, our proposed method, generates pairs without relying on human-curated data, enabling a more autonomous approach.

\begin{figure*}[h]
    \centering
    \subfloat[From English into Zulu]{
        \includegraphics[width=0.5\textwidth]{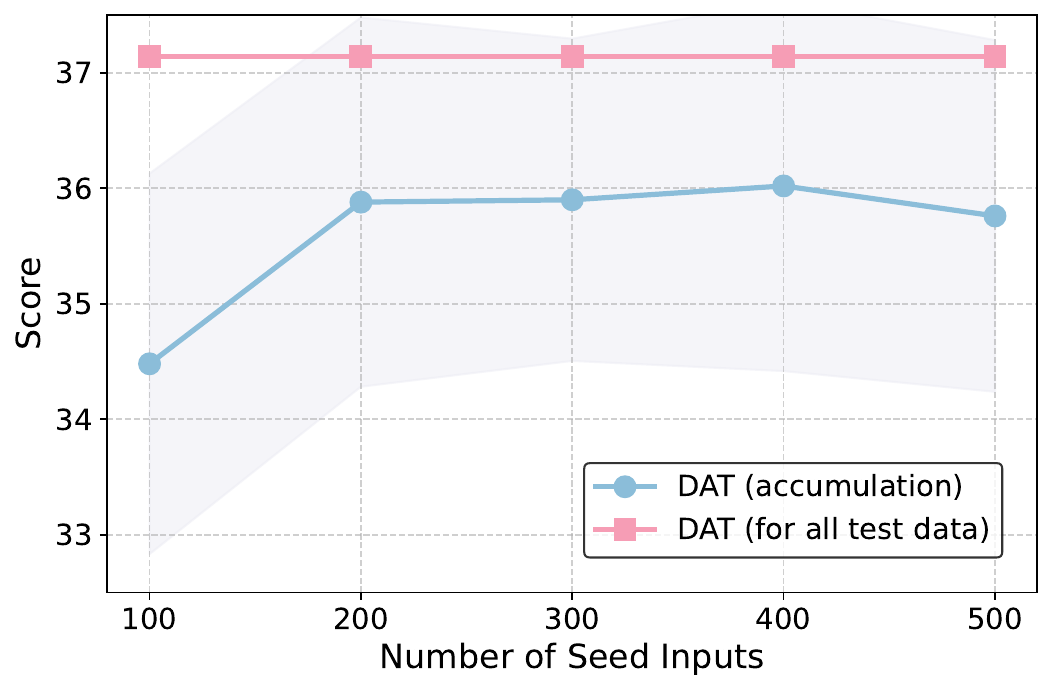}
        \label{fig:image1}
    }
    \subfloat[From English into Swahili]{
        \includegraphics[width=0.5\textwidth]{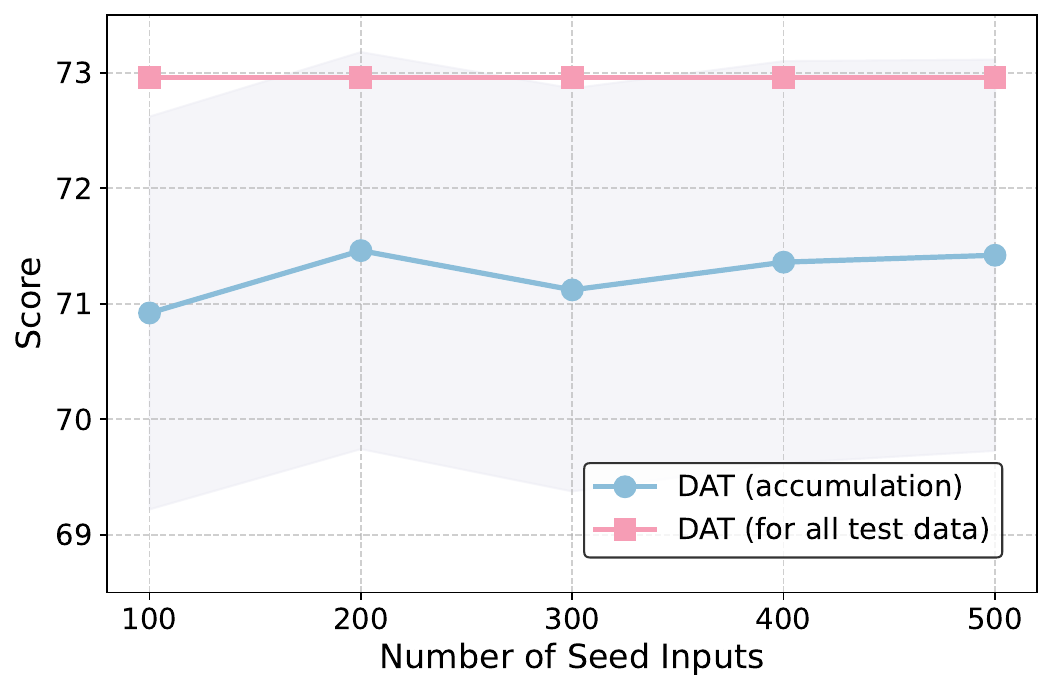}
        \label{fig:image2}
    }
    \caption{This experiment evaluates the accumulation setting when translating English into Zulu and Swahili using Llama-3.1-70B. The x-axis represents the number of test inputs used to construct the demonstration pool, while performance is assessed on a fixed set of 512 test samples, independent of the x value. To quantify variability, we further divide the test data into five distinct subsets and report the standard deviation.}
    \label{fig:accum}
\end{figure*}
\paragraph{Source-side of Pair}
Relevance and Uniformity exhibit the highest values in DAT and the lowest in the Fixed set.
This disparity arises from the fact that both DAT and Retrieval dynamically generate or procure source sentences based on the user query, whereas the Fixed set operates independently of such adaptation.
Meanwhile, Relevance and Uniformity appear to be interdependent, with an increase in Relevance generally leading to an increase in Uniformity.
Though finding an appropriate balance between the two is important, Relevance is generally considered the higher-priority measure, as in-context examples that are diverse but unrelated to the user query are unlikely to be beneficial for translation.
This is also supported by the DAT with the highest Relevance achieving a higher COMET score than other baselines.

\paragraph{Source-Target Pair}
While Relevance and Uniformity—focusing only on the source side of in-context examples—are important metrics, the overall quality of the pair itself is also a crucial factor.
We evaluate this quality using the reference-free COMET score, reported in the Quality column.
Despite the undeniable superiority of human-curated pairs in the Fixed set in terms of quality, our experimental results demonstrate that they yield lower translation performance compared to other baselines. 
This phenomenon can be attributed to their insufficient Relevance, which hinders their direct contribution to user query translation.
This finding suggests that, rather than employing high-quality pairs that fail to ensure relevance to the user query, a more effective approach would be to generate or retrieve sentences with guaranteed relevance and then artificially construct their corresponding target sentences for use as in-context examples.

\subsection{Ablation Study on MMR Filtering}
\begin{table}[h]
\small
\begin{tabular}{@{}c|cc|ccc@{}}
\toprule
\textbf{Method} & \textbf{m} & \textbf{k} & \textbf{Khmer} & \textbf{Pashto} & \textbf{Swahili} \\ \midrule
No Filtering$_{4}$    & 4          & 4          & 63.8           & 54.2            & 61.9             \\
No Filtering$_{10}$    & 10         & 10         & 63.4           & 53.6            & 61.7             \\ \midrule
DAT             & 10         & 4          & \textbf{64.4}  & \textbf{54.6}   & \textbf{62.3}    \\ \bottomrule
\end{tabular}
\caption{
This result illustrates the impact of $m$ and $k$. 
$m$ denotes the number of source sentences generated initially, whereas $k$ signifies the final number of demonstrations after filtering process. 
This means that when $m$ and $k$ are the same, the filtering process is not applied. 
The best performances are in \textbf{bold}.
}
\label{table:ablation_filtering}
\end{table}
Our method applies a filtering process to select only $k$ sentences with high relevance and diversity from the initially generated $m$ sentences.
Then, we perform translation on the remaining $k$ sentences to generate synthetic pairs by utilizing the LLM.
To verify the validity of process, we analyze the COMET scores when translating English into three low-resource languages in Table~\ref{table:ablation_filtering}.
It is noteworthy that No Filtering$_{10}$ showed inferior performance compared to No Filtering$_{4}$, despite using more in-context examples.
This means that the additional $6$ demonstrations acted as noise, degrading the translation quality.
Furthermore, the results demonstrate that our approach—generating $m$ source sentences with $m$ > $k$ and then applying filtering—achieves superior performance compared to other methods.
This validates the effectiveness of the filtering process and implies that using a small number of carefully selected demonstrations, which are filtered to better assist the user query, can lead to performance gains than utilizing $10$ unfiltered demonstrations.

\subsection{Accumulation Setting}
\label{sec:accum}
\paragraph{Experimental Configuration}
In previous experiments, DAT demonstrated superior performance compared to baseline methods. 
However, generating in-context examples for every test input can be computationally expensive. 
To address this, we explore a more cost-efficient approach in this experiment. 
Specifically, we incrementally construct a demonstration pool by accumulating pairs generated for a subset of test inputs. 
Subsequently, we select in-context examples for translation based on their R-BM25 scores, thereby enhancing efficiency.

We partition the 1,012 test samples from the Flores dataset into two subsets of 500 and 512 instances. 
The first subset of 500 samples is designated as seed data, which we use to generate and accumulate pairs, thereby constructing the demonstration pool. 
Performance is then evaluated exclusively on the remaining 512 samples. 
Additionally, to investigate the impact of increasing seed data on performance, we progressively increase the number of seed data and assess the resulting improvements.

\begin{table*}[h]
\centering
\tiny
\begin{tabular}{@{}c|c|c|cc|c@{}}
\toprule
\textbf{Test Data}                                                                                                                                                             & \textbf{Reference}                                                                                                                                                                                                    & \textbf{Method} & \multicolumn{2}{c|}{\textbf{In-context Example}}                                                                                                                                                                                                                                                                                          & \textbf{Hypothesis}                                                                                                                                                           \\ \midrule
\multirow{2}{*}{\begin{tabular}[c]{@{}c@{}}Prides are made \\ up of one to \\ three related \\ adult males, along with \\ as many as thirty \\ females and cubs.\end{tabular}} & \multirow{2}{*}{\begin{tabular}[c]{@{}c@{}}Makundi ya \colorbox{blue!15}{simba} \\ yanatengenezwa \colorbox{gray!20}{na} \\ \colorbox{gray!20}{waume} kati ya \\ mmoja na \colorbox{blue!15}{watatu} \\ wazima wanaohusiana, \\ \colorbox{gray!20}{pamoja na} wengi wa kike \\ kama \colorbox{blue!15}{thelathini} \colorbox{gray!20}{na} mashibli.\end{tabular}} & Zero-shot       & \multicolumn{2}{c|}{}                                                                                                                                                                                                                                                                                                                     & \begin{tabular}[c]{@{}c@{}}Vichwa vimeundwa \colorbox{gray!20}{na} \\ \colorbox{gray!20}{wanaume} wazima \\ 1 hadi 3 wenye uhusiano, \\ \colorbox{gray!20}{pamoja na} wanawake \\ hadi 30 \colorbox{gray!20}{na} watoto.\end{tabular}                         \\ \cmidrule(l){3-6} 
                                                                                                                                                                               &                                                                                                                                                                                                                       & DAT             & \begin{tabular}[c]{@{}c@{}}A pride of lions \\ can consist of \\ up to thirty females \\ and cubs, accompanied by \\ one to three adult males.\end{tabular} & \begin{tabular}[c]{@{}c@{}}Kundi la \colorbox{blue!15}{simba} linaweza \\ kuwa na hadi wanawake \\ \colorbox{blue!15}{thelathini} na watoto, \\ wanaoambatana na wanaume \\ wazima wawili hadi \colorbox{blue!15}{watatu}.\end{tabular} & \begin{tabular}[c]{@{}c@{}}Kundi la \colorbox{blue!15}{simba} huundwa \\ \colorbox{gray!20}{na wanaume} wazima \\ wawili hadi \colorbox{blue!15}{watatu} wenye \\ uhusiano, \colorbox{gray!20}{pamoja na} wanawake \\ hadi \colorbox{blue!15}{thelathini} \colorbox{gray!20}{na} watoto.\end{tabular} \\ \bottomrule
\end{tabular}
\caption{
This table shows a surface-level analysis when translating English to Swahili.
\fcolorbox{gray!20}{gray!20}{\raisebox{0pt}[6pt][0pt]{Gray}} indicates cases where both Zero-shot and DAT produced the correct terms. \fcolorbox{blue!15}{blue!15}{\raisebox{0pt}[6pt][0pt]{Blue}} represents cases where only DAT generated the correct terms, demonstrating that DAT benefited from the self-generated in-context examples.}
\label{table:surface_analysis}
\end{table*}
\paragraph{Result}
Figure~\ref{fig:accum} compares the COMET score of DAT, which performs few-shot learning by generating pairs for all test input, and DAT (Accumulation), which leverages a demonstration pool consisting of progressively accumulated pairs after a certain point.
In both Zulu and Swahili, an increasing number of seed inputs exhibits a general trend of performance improvement. 
However, when employing a demonstration pool constructed from up to 500 data points, this approach does not fully reach the performance level of the method that generates pairs dynamically for each test instance. 
Nevertheless, we hypothesize that as more seed input is incorporated, a larger and more diverse demonstration pool can be constructed, ultimately enabling high-quality translations without the need for on-the-fly pair generation.
Future research should explore optimal strategies for constructing and expanding such a pool, ensuring robust performance even in low-resource scenarios.

\subsection{Surface-level Analysis}
Table~\ref{table:surface_analysis} illustrates the impact of in-context examples generated via DAT on translation outcomes, offering empirical evidence of their effectiveness. 
In this example, DAT successfully produces precise lexical choices that the Zero-shot approach fails to achieve. 
This is due to the alignment of generated in-context examples with the user query, which offers crucial contextual cues for precise translation.

\section{Discussions}
\paragraph{Hybrid Approach}
We employed an LLM to generate the target-side of in-context examples. While a model fine-tuned for a specific language pair—such as a traditional neural machine translation system—could produce higher-fidelity pairs, our focus is on improving translation quality purely through the inherent capabilities of the LLM. We therefore leave the exploration of such approaches as a promising direction for future work.

\paragraph{Reuse as a Training Dataset}
In Section~\ref{sec:accum}, we explored a setting where the generated pairs are accumulated, allowing for retrieval and reuse.
If the demonstration pool becomes large enough, it can serve as a training dataset for developing a translation-specialized model. Therefore, leveraging LLMs to generate translation pairs in low-resource scenarios is a crucial research direction, both for in-context learning and fine-tuning.

\paragraph{Post Editing}
Another line of research in translation using LLMs is post-editing, which focuses on refining the initial translation~\citep{raunak-etal-2023-leveraging}.
Our research can be effectively combined with this method. 
In post-editing approaches, the quality of the initial translation is crucial. 
Applying DAT to generate the initial translation in scenarios where no in-context examples are available and then refining it presents a highly promising translation strategy.

\section{Conclusions}
In this paper, we explore an interesting research direction that leverages only LLMs to generate source-target pairs.
Accordingly, we propose a simple yet effective method, DAT, which utilizes two priors and MMR to generate in-context examples.
Experiments proved that DAT achieves superior translation quality for low-resource languages compared to baselines, without relying on any human-created resources.
This highlights the potential of our method in scenarios where a demonstration pool is unavailable or fails to provide relevant translation examples.
Furthermore, we investigated cases where fixed human pairs significantly underperform compared to zero-shot translation and explored the potential of an accumulation setting as a cost-efficient alternative.

\section{Limitations}
\paragraph{Translation from a Low-resource Language}
In this work, we focus on translating English into other low-resource languages. This is because most LLMs are primarily trained on English, allowing them to generate high-quality in-context examples on the source side. 
In contrast, translating in the opposite direction would heavily rely on the LLM’s monolingual generation capabilities for low-resource languages. 
Therefore, this setting is not explored in this paper and remains an important challenge for future research.

\paragraph{Open-source LLMs}
Most open-source LLMs are primarily trained with a strong emphasis on English, limiting their effectiveness in handling a diverse set of languages. 
Given the scarcity of open-source models that offer robust multilingual support, we conduct our experiments using Llama-3.1, an open-source LLM designed to support a wide range of languages, including English.

\paragraph{Accumulation Setting}
In DAT with accumulation, the performance did not fully reach that of generating pairs for all test data.
We aimed to create a larger demonstration pool to explore the point at which performance reaches that of generating pairs for all test data.
However, this investigation could not be performed due to the high computational cost and is left for future work.

\section{Ethical Considerations}
We conducted our experiments using publicly available datasets and models, and the datasets do not involve any ethical concerns.

\section{Acknowledgments}
This work was supported by SAMSUNG Research, Samsung Electronics Co., Ltd., Institute for Information \& communications Technology Planning \& Evaluation(IITP) grant funded by the Korea government(MSIT) (RS-2019-II190075, Artificial Intelligence Graduate School Program(KAIST)), and the National Research Foundation of Korea(NRF) grant funded by the Korea government(MSIT) (No. RS-2025-00555621).

\bibliography{custom}

\begin{thebibliography}{51}
\providecommand{\natexlab}[1]{#1}

\bibitem[{Agrawal et~al.(2023)Agrawal, Zhou, Lewis, Zettlemoyer, and Ghazvininejad}]{rbm25}
Sweta Agrawal, Chunting Zhou, Mike Lewis, Luke Zettlemoyer, and Marjan Ghazvininejad. 2023.
\newblock \href {https://doi.org/10.18653/v1/2023.findings-acl.564} {In-context examples selection for machine translation}.
\newblock In \emph{Findings of the Association for Computational Linguistics: ACL 2023}, pages 8857--8873, Toronto, Canada. Association for Computational Linguistics.

\bibitem[{Bouthors et~al.(2024)Bouthors, Crego, and Yvon}]{bouthors-etal-2024-retrieving}
Maxime Bouthors, Josep Crego, and Fran{\c{c}}ois Yvon. 2024.
\newblock \href {https://doi.org/10.18653/v1/2024.findings-naacl.190} {Retrieving examples from memory for retrieval augmented neural machine translation: A systematic comparison}.
\newblock In \emph{Findings of the Association for Computational Linguistics: NAACL 2024}, pages 3022--3039, Mexico City, Mexico. Association for Computational Linguistics.

\bibitem[{Brown et~al.(2020)Brown, Mann, Ryder, Subbiah, Kaplan, Dhariwal, Neelakantan, Shyam, Sastry, Askell, Agarwal, Herbert-Voss, Krueger, Henighan, Child, Ramesh, Ziegler, Wu, Winter, Hesse, Chen, Sigler, Litwin, Gray, Chess, Clark, Berner, McCandlish, Radford, Sutskever, and Amodei}]{icl}
Tom Brown, Benjamin Mann, Nick Ryder, Melanie Subbiah, Jared~D Kaplan, Prafulla Dhariwal, Arvind Neelakantan, Pranav Shyam, Girish Sastry, Amanda Askell, Sandhini Agarwal, Ariel Herbert-Voss, Gretchen Krueger, Tom Henighan, Rewon Child, Aditya Ramesh, Daniel Ziegler, Jeffrey Wu, Clemens Winter, Chris Hesse, Mark Chen, Eric Sigler, Mateusz Litwin, Scott Gray, Benjamin Chess, Jack Clark, Christopher Berner, Sam McCandlish, Alec Radford, Ilya Sutskever, and Dario Amodei. 2020.
\newblock \href {https://proceedings.neurips.cc/paper_files/paper/2020/file/1457c0d6bfcb4967418bfb8ac142f64a-Paper.pdf} {Language models are few-shot learners}.
\newblock In \emph{Advances in Neural Information Processing Systems}, volume~33, pages 1877--1901. Curran Associates, Inc.

\bibitem[{Carbonell and Goldstein(1998)}]{maximal_marginal_relevance}
Jaime Carbonell and Jade Goldstein. 1998.
\newblock \href {https://doi.org/10.1145/290941.291025} {The use of mmr, diversity-based reranking for reordering documents and producing summaries}.
\newblock In \emph{Proceedings of the 21st Annual International ACM SIGIR Conference on Research and Development in Information Retrieval}, SIGIR '98, page 335–336, New York, NY, USA. Association for Computing Machinery.

\bibitem[{Chan et~al.(2022)Chan, Santoro, Lampinen, Wang, Singh, Richemond, McClelland, and Hill}]{distproperty}
Stephanie C.~Y. Chan, Adam Santoro, Andrew~K. Lampinen, Jane~X. Wang, Aaditya~K. Singh, Pierre~H. Richemond, James~L. McClelland, and Felix Hill. 2022.
\newblock \href {http://papers.nips.cc/paper\_files/paper/2022/hash/77c6ccacfd9962e2307fc64680fc5ace-Abstract-Conference.html} {Data distributional properties drive emergent in-context learning in transformers}.
\newblock In \emph{Advances in Neural Information Processing Systems 35: Annual Conference on Neural Information Processing Systems 2022, NeurIPS 2022, New Orleans, LA, USA, November 28 - December 9, 2022}.

\bibitem[{Chen et~al.(2023)Chen, Wu, Chen, and Chen}]{self-icl}
Wei{-}Lin Chen, Cheng{-}Kuang Wu, Yun{-}Nung Chen, and Hsin{-}Hsi Chen. 2023.
\newblock \href {https://doi.org/10.18653/V1/2023.EMNLP-MAIN.968} {Self-icl: Zero-shot in-context learning with self-generated demonstrations}.
\newblock In \emph{Proceedings of the 2023 Conference on Empirical Methods in Natural Language Processing, {EMNLP} 2023, Singapore, December 6-10, 2023}, pages 15651--15662. Association for Computational Linguistics.

\bibitem[{Cheng et~al.(2023)Cheng, Gan, Yang, Wang, Wang, Boyd-Graber, and Wang}]{prompting-reliable}
Silei Cheng, Zhe Gan, Zhengyuan Yang, Shuohang Wang, Jianfeng Wang, Jordan Boyd-Graber, and Lijuan Wang. 2023.
\newblock \href {https://www.microsoft.com/en-us/research/publication/prompting-gpt-3-to-be-reliable/} {Prompting gpt-3 to be reliable}.
\newblock In \emph{International Conference on Learning Representations (ICLR 23)}.

\bibitem[{Cheng et~al.(2022)Cheng, Gao, Liu, Zhao, and Yan}]{cheng-etal-2022-neural}
Xin Cheng, Shen Gao, Lemao Liu, Dongyan Zhao, and Rui Yan. 2022.
\newblock \href {https://doi.org/10.18653/v1/2022.emnlp-main.235} {Neural machine translation with contrastive translation memories}.
\newblock In \emph{Proceedings of the 2022 Conference on Empirical Methods in Natural Language Processing}, pages 3591--3601, Abu Dhabi, United Arab Emirates. Association for Computational Linguistics.

\bibitem[{Chowdhery et~al.(2023)Chowdhery, Narang, Devlin, Bosma, Mishra, Roberts, Barham, Chung, Sutton, Gehrmann, Schuh, Shi, Tsvyashchenko, Maynez, Rao, Barnes, Tay, Shazeer, Prabhakaran, Reif, Du, Hutchinson, Pope, Bradbury, Austin, Isard, Gur-Ari, Yin, Duke, Levskaya, Ghemawat, Dev, Michalewski, Garcia, Misra, Robinson, Fedus, Zhou, Ippolito, Luan, Lim, Zoph, Spiridonov, Sepassi, Dohan, Agrawal, Omernick, Dai, Pillai, Pellat, Lewkowycz, Moreira, Child, Polozov, Lee, Zhou, Wang, Saeta, Diaz, Firat, Catasta, Wei, Meier-Hellstern, Eck, Dean, Petrov, and Fiedel}]{palm}
Aakanksha Chowdhery, Sharan Narang, Jacob Devlin, Maarten Bosma, Gaurav Mishra, Adam Roberts, Paul Barham, Hyung~Won Chung, Charles Sutton, Sebastian Gehrmann, Parker Schuh, Kensen Shi, Sashank Tsvyashchenko, Joshua Maynez, Abhishek Rao, Parker Barnes, Yi~Tay, Noam Shazeer, Vinodkumar Prabhakaran, Emily Reif, Nan Du, Ben Hutchinson, Reiner Pope, James Bradbury, Jacob Austin, Michael Isard, Guy Gur-Ari, Pengcheng Yin, Toju Duke, Anselm Levskaya, Sanjay Ghemawat, Sunipa Dev, Henryk Michalewski, Xavier Garcia, Vedant Misra, Kevin Robinson, Liam Fedus, Denny Zhou, Daphne Ippolito, David Luan, Hyeontaek Lim, Barret Zoph, Alexander Spiridonov, Ryan Sepassi, David Dohan, Shivani Agrawal, Mark Omernick, Andrew~M. Dai, Thanumalayan~Sankaranarayana Pillai, Marie Pellat, Aitor Lewkowycz, Erica Moreira, Rewon Child, Oleksandr Polozov, Katherine Lee, Zongwei Zhou, Xuezhi Wang, Brennan Saeta, Mark Diaz, Orhan Firat, Michele Catasta, Jason Wei, Kathy Meier-Hellstern, Douglas Eck, Jeff Dean, Slav Petrov, and Noah Fiedel.
  2023.
\newblock Palm: scaling language modeling with pathways.
\newblock \emph{J. Mach. Learn. Res.}, 24(1).

\bibitem[{Dubey et~al.(2024)Dubey, Jauhri, Pandey, Kadian, Al-Dahle, Letman, Mathur, Schelten, Yang, Fan, Goyal, Hartshorn, Yang, Mitra, Sravankumar, Korenev, Hinsvark, Rao, Zhang, Rodriguez, Gregerson, Spataru, Roziere, Biron, Tang, Chern, Caucheteux, Nayak, Bi, Marra, McConnell, Keller, Touret, Wu, Wong, Ferrer, Nikolaidis, Allonsius, Song, Pintz, Livshits, Esiobu, Choudhary, Mahajan, Garcia-Olano, Perino, Hupkes, Lakomkin, AlBadawy, Lobanova, Dinan, Smith, Radenovic, Zhang, Synnaeve, Lee, Anderson, Nail, Mialon, Pang, Cucurell, Nguyen, Korevaar, Xu, Touvron, Zarov, Ibarra, Kloumann, Misra, Evtimov, Copet, Lee, Geffert, Vranes, Park, Mahadeokar, Shah, van~der Linde, Billock, Hong, Lee, Fu, Chi, Huang, Liu, Wang, Yu, Bitton, Spisak, Park, Rocca, Johnstun, Saxe, Jia, Alwala, Upasani, Plawiak, Li, Heafield, Stone, El-Arini, Iyer, Malik, Chiu, Bhalla, Rantala-Yeary, van~der Maaten, Chen, Tan, Jenkins, Martin, Madaan, Malo, Blecher, Landzaat, de~Oliveira, Muzzi, Pasupuleti, Singh, Paluri, Kardas, Oldham, Rita,
  Pavlova, Kambadur, Lewis, Si, Singh, Hassan, Goyal, Torabi, Bashlykov, Bogoychev, Chatterji, Duchenne, Çelebi, Alrassy, Zhang, Li, Vasic, Weng, Bhargava, Dubal, Krishnan, Koura, Xu, He, Dong, Srinivasan, Ganapathy, Calderer, Cabral, Stojnic, Raileanu, Girdhar, Patel, Sauvestre, Polidoro, Sumbaly, Taylor, Silva, Hou, Wang, Hosseini, Chennabasappa, Singh, Bell, Kim, Edunov, Nie, Narang, Raparthy, Shen, Wan, Bhosale, Zhang, Vandenhende, Batra, Whitman, Sootla, Collot, Gururangan, Borodinsky, Herman, Fowler, Sheasha, Georgiou, Scialom, Speckbacher, Mihaylov, Xiao, Karn, Goswami, Gupta, Ramanathan, Kerkez, Gonguet, Do, Vogeti, Petrovic, Chu, Xiong, Fu, Meers, Martinet, Wang, Tan, Xie, Jia, Wang, Goldschlag, Gaur, Babaei, Wen, Song, Zhang, Li, Mao, Coudert, Yan, Chen, Papakipos, Singh, Grattafiori, Jain, Kelsey, Shajnfeld, Gangidi, Victoria, Goldstand, Menon, Sharma, Boesenberg, Vaughan, Baevski, Feinstein, Kallet, Sangani, Yunus, Lupu, Alvarado, Caples, Gu, Ho, Poulton, Ryan, Ramchandani, Franco, Saraf,
  Chowdhury, Gabriel, Bharambe, Eisenman, Yazdan, James, Maurer, Leonhardi, Huang, Loyd, Paola, Paranjape, Liu, Wu, Ni, Hancock, Wasti, Spence, Stojkovic, Gamido, Montalvo, Parker, Burton, Mejia, Wang, Kim, Zhou, Hu, Chu, Cai, Tindal, Feichtenhofer, Civin, Beaty, Kreymer, Li, Wyatt, Adkins, Xu, Testuggine, David, Parikh, Liskovich, Foss, Wang, Le, Holland, Dowling, Jamil, Montgomery, Presani, Hahn, Wood, Brinkman, Arcaute, Dunbar, Smothers, Sun, Kreuk, Tian, Ozgenel, Caggioni, Guzmán, Kanayet, Seide, Florez, Schwarz, Badeer, Swee, Halpern, Thattai, Herman, Sizov, Guangyi, Zhang, Lakshminarayanan, Shojanazeri, Zou, Wang, Zha, Habeeb, Rudolph, Suk, Aspegren, Goldman, Damlaj, Molybog, Tufanov, Veliche, Gat, Weissman, Geboski, Kohli, Asher, Gaya, Marcus, Tang, Chan, Zhen, Reizenstein, Teboul, Zhong, Jin, Yang, Cummings, Carvill, Shepard, McPhie, Torres, Ginsburg, Wang, Wu, U, Saxena, Prasad, Khandelwal, Zand, Matosich, Veeraraghavan, Michelena, Li, Huang, Chawla, Lakhotia, Huang, Chen, Garg, A, Silva, Bell,
  Zhang, Guo, Yu, Moshkovich, Wehrstedt, Khabsa, Avalani, Bhatt, Tsimpoukelli, Mankus, Hasson, Lennie, Reso, Groshev, Naumov, Lathi, Keneally, Seltzer, Valko, Restrepo, Patel, Vyatskov, Samvelyan, Clark, Macey, Wang, Hermoso, Metanat, Rastegari, Bansal, Santhanam, Parks, White, Bawa, Singhal, Egebo, Usunier, Laptev, Dong, Zhang, Cheng, Chernoguz, Hart, Salpekar, Kalinli, Kent, Parekh, Saab, Balaji, Rittner, Bontrager, Roux, Dollar, Zvyagina, Ratanchandani, Yuvraj, Liang, Alao, Rodriguez, Ayub, Murthy, Nayani, Mitra, Li, Hogan, Battey, Wang, Maheswari, Howes, Rinott, Bondu, Datta, Chugh, Hunt, Dhillon, Sidorov, Pan, Verma, Yamamoto, Ramaswamy, Lindsay, Lindsay, Feng, Lin, Zha, Shankar, Zhang, Zhang, Wang, Agarwal, Sajuyigbe, Chintala, Max, Chen, Kehoe, Satterfield, Govindaprasad, Gupta, Cho, Virk, Subramanian, Choudhury, Goldman, Remez, Glaser, Best, Kohler, Robinson, Li, Zhang, Matthews, Chou, Shaked, Vontimitta, Ajayi, Montanez, Mohan, Kumar, Mangla, Albiero, Ionescu, Poenaru, Mihailescu, Ivanov, Li, Wang,
  Jiang, Bouaziz, Constable, Tang, Wang, Wu, Wang, Xia, Wu, Gao, Chen, Hu, Jia, Qi, Li, Zhang, Zhang, Adi, Nam, Yu, Wang, Hao, Qian, He, Rait, DeVito, Rosnbrick, Wen, Yang, and Zhao}]{llama3.1}
Abhimanyu Dubey, Abhinav Jauhri, Abhinav Pandey, Abhishek Kadian, Ahmad Al-Dahle, Aiesha Letman, Akhil Mathur, Alan Schelten, Amy Yang, Angela Fan, Anirudh Goyal, Anthony Hartshorn, Aobo Yang, Archi Mitra, Archie Sravankumar, Artem Korenev, Arthur Hinsvark, Arun Rao, Aston Zhang, Aurelien Rodriguez, Austen Gregerson, Ava Spataru, Baptiste Roziere, Bethany Biron, Binh Tang, Bobbie Chern, Charlotte Caucheteux, Chaya Nayak, Chloe Bi, Chris Marra, Chris McConnell, Christian Keller, Christophe Touret, Chunyang Wu, Corinne Wong, Cristian~Canton Ferrer, Cyrus Nikolaidis, Damien Allonsius, Daniel Song, Danielle Pintz, Danny Livshits, David Esiobu, Dhruv Choudhary, Dhruv Mahajan, Diego Garcia-Olano, Diego Perino, Dieuwke Hupkes, Egor Lakomkin, Ehab AlBadawy, Elina Lobanova, Emily Dinan, Eric~Michael Smith, Filip Radenovic, Frank Zhang, Gabriel Synnaeve, Gabrielle Lee, Georgia~Lewis Anderson, Graeme Nail, Gregoire Mialon, Guan Pang, Guillem Cucurell, Hailey Nguyen, Hannah Korevaar, Hu~Xu, Hugo Touvron, Iliyan Zarov,
  Imanol~Arrieta Ibarra, Isabel Kloumann, Ishan Misra, Ivan Evtimov, Jade Copet, Jaewon Lee, Jan Geffert, Jana Vranes, Jason Park, Jay Mahadeokar, Jeet Shah, Jelmer van~der Linde, Jennifer Billock, Jenny Hong, Jenya Lee, Jeremy Fu, Jianfeng Chi, Jianyu Huang, Jiawen Liu, Jie Wang, Jiecao Yu, Joanna Bitton, Joe Spisak, Jongsoo Park, Joseph Rocca, Joshua Johnstun, Joshua Saxe, Junteng Jia, Kalyan~Vasuden Alwala, Kartikeya Upasani, Kate Plawiak, Ke~Li, Kenneth Heafield, Kevin Stone, Khalid El-Arini, Krithika Iyer, Kshitiz Malik, Kuenley Chiu, Kunal Bhalla, Lauren Rantala-Yeary, Laurens van~der Maaten, Lawrence Chen, Liang Tan, Liz Jenkins, Louis Martin, Lovish Madaan, Lubo Malo, Lukas Blecher, Lukas Landzaat, Luke de~Oliveira, Madeline Muzzi, Mahesh Pasupuleti, Mannat Singh, Manohar Paluri, Marcin Kardas, Mathew Oldham, Mathieu Rita, Maya Pavlova, Melanie Kambadur, Mike Lewis, Min Si, Mitesh~Kumar Singh, Mona Hassan, Naman Goyal, Narjes Torabi, Nikolay Bashlykov, Nikolay Bogoychev, Niladri Chatterji, Olivier
  Duchenne, Onur Çelebi, Patrick Alrassy, Pengchuan Zhang, Pengwei Li, Petar Vasic, Peter Weng, Prajjwal Bhargava, Pratik Dubal, Praveen Krishnan, Punit~Singh Koura, Puxin Xu, Qing He, Qingxiao Dong, Ragavan Srinivasan, Raj Ganapathy, Ramon Calderer, Ricardo~Silveira Cabral, Robert Stojnic, Roberta Raileanu, Rohit Girdhar, Rohit Patel, Romain Sauvestre, Ronnie Polidoro, Roshan Sumbaly, Ross Taylor, Ruan Silva, Rui Hou, Rui Wang, Saghar Hosseini, Sahana Chennabasappa, Sanjay Singh, Sean Bell, Seohyun~Sonia Kim, Sergey Edunov, Shaoliang Nie, Sharan Narang, Sharath Raparthy, Sheng Shen, Shengye Wan, Shruti Bhosale, Shun Zhang, Simon Vandenhende, Soumya Batra, Spencer Whitman, Sten Sootla, Stephane Collot, Suchin Gururangan, Sydney Borodinsky, Tamar Herman, Tara Fowler, Tarek Sheasha, Thomas Georgiou, Thomas Scialom, Tobias Speckbacher, Todor Mihaylov, Tong Xiao, Ujjwal Karn, Vedanuj Goswami, Vibhor Gupta, Vignesh Ramanathan, Viktor Kerkez, Vincent Gonguet, Virginie Do, Vish Vogeti, Vladan Petrovic, Weiwei Chu,
  Wenhan Xiong, Wenyin Fu, Whitney Meers, Xavier Martinet, Xiaodong Wang, Xiaoqing~Ellen Tan, Xinfeng Xie, Xuchao Jia, Xuewei Wang, Yaelle Goldschlag, Yashesh Gaur, Yasmine Babaei, Yi~Wen, Yiwen Song, Yuchen Zhang, Yue Li, Yuning Mao, Zacharie~Delpierre Coudert, Zheng Yan, Zhengxing Chen, Zoe Papakipos, Aaditya Singh, Aaron Grattafiori, Abha Jain, Adam Kelsey, Adam Shajnfeld, Adithya Gangidi, Adolfo Victoria, Ahuva Goldstand, Ajay Menon, Ajay Sharma, Alex Boesenberg, Alex Vaughan, Alexei Baevski, Allie Feinstein, Amanda Kallet, Amit Sangani, Anam Yunus, Andrei Lupu, Andres Alvarado, Andrew Caples, Andrew Gu, Andrew Ho, Andrew Poulton, Andrew Ryan, Ankit Ramchandani, Annie Franco, Aparajita Saraf, Arkabandhu Chowdhury, Ashley Gabriel, Ashwin Bharambe, Assaf Eisenman, Azadeh Yazdan, Beau James, Ben Maurer, Benjamin Leonhardi, Bernie Huang, Beth Loyd, Beto~De Paola, Bhargavi Paranjape, Bing Liu, Bo~Wu, Boyu Ni, Braden Hancock, Bram Wasti, Brandon Spence, Brani Stojkovic, Brian Gamido, Britt Montalvo, Carl
  Parker, Carly Burton, Catalina Mejia, Changhan Wang, Changkyu Kim, Chao Zhou, Chester Hu, Ching-Hsiang Chu, Chris Cai, Chris Tindal, Christoph Feichtenhofer, Damon Civin, Dana Beaty, Daniel Kreymer, Daniel Li, Danny Wyatt, David Adkins, David Xu, Davide Testuggine, Delia David, Devi Parikh, Diana Liskovich, Didem Foss, Dingkang Wang, Duc Le, Dustin Holland, Edward Dowling, Eissa Jamil, Elaine Montgomery, Eleonora Presani, Emily Hahn, Emily Wood, Erik Brinkman, Esteban Arcaute, Evan Dunbar, Evan Smothers, Fei Sun, Felix Kreuk, Feng Tian, Firat Ozgenel, Francesco Caggioni, Francisco Guzmán, Frank Kanayet, Frank Seide, Gabriela~Medina Florez, Gabriella Schwarz, Gada Badeer, Georgia Swee, Gil Halpern, Govind Thattai, Grant Herman, Grigory Sizov, Guangyi, Zhang, Guna Lakshminarayanan, Hamid Shojanazeri, Han Zou, Hannah Wang, Hanwen Zha, Haroun Habeeb, Harrison Rudolph, Helen Suk, Henry Aspegren, Hunter Goldman, Ibrahim Damlaj, Igor Molybog, Igor Tufanov, Irina-Elena Veliche, Itai Gat, Jake Weissman, James
  Geboski, James Kohli, Japhet Asher, Jean-Baptiste Gaya, Jeff Marcus, Jeff Tang, Jennifer Chan, Jenny Zhen, Jeremy Reizenstein, Jeremy Teboul, Jessica Zhong, Jian Jin, Jingyi Yang, Joe Cummings, Jon Carvill, Jon Shepard, Jonathan McPhie, Jonathan Torres, Josh Ginsburg, Junjie Wang, Kai Wu, Kam~Hou U, Karan Saxena, Karthik Prasad, Kartikay Khandelwal, Katayoun Zand, Kathy Matosich, Kaushik Veeraraghavan, Kelly Michelena, Keqian Li, Kun Huang, Kunal Chawla, Kushal Lakhotia, Kyle Huang, Lailin Chen, Lakshya Garg, Lavender A, Leandro Silva, Lee Bell, Lei Zhang, Liangpeng Guo, Licheng Yu, Liron Moshkovich, Luca Wehrstedt, Madian Khabsa, Manav Avalani, Manish Bhatt, Maria Tsimpoukelli, Martynas Mankus, Matan Hasson, Matthew Lennie, Matthias Reso, Maxim Groshev, Maxim Naumov, Maya Lathi, Meghan Keneally, Michael~L. Seltzer, Michal Valko, Michelle Restrepo, Mihir Patel, Mik Vyatskov, Mikayel Samvelyan, Mike Clark, Mike Macey, Mike Wang, Miquel~Jubert Hermoso, Mo~Metanat, Mohammad Rastegari, Munish Bansal, Nandhini
  Santhanam, Natascha Parks, Natasha White, Navyata Bawa, Nayan Singhal, Nick Egebo, Nicolas Usunier, Nikolay~Pavlovich Laptev, Ning Dong, Ning Zhang, Norman Cheng, Oleg Chernoguz, Olivia Hart, Omkar Salpekar, Ozlem Kalinli, Parkin Kent, Parth Parekh, Paul Saab, Pavan Balaji, Pedro Rittner, Philip Bontrager, Pierre Roux, Piotr Dollar, Polina Zvyagina, Prashant Ratanchandani, Pritish Yuvraj, Qian Liang, Rachad Alao, Rachel Rodriguez, Rafi Ayub, Raghotham Murthy, Raghu Nayani, Rahul Mitra, Raymond Li, Rebekkah Hogan, Robin Battey, Rocky Wang, Rohan Maheswari, Russ Howes, Ruty Rinott, Sai~Jayesh Bondu, Samyak Datta, Sara Chugh, Sara Hunt, Sargun Dhillon, Sasha Sidorov, Satadru Pan, Saurabh Verma, Seiji Yamamoto, Sharadh Ramaswamy, Shaun Lindsay, Shaun Lindsay, Sheng Feng, Shenghao Lin, Shengxin~Cindy Zha, Shiva Shankar, Shuqiang Zhang, Shuqiang Zhang, Sinong Wang, Sneha Agarwal, Soji Sajuyigbe, Soumith Chintala, Stephanie Max, Stephen Chen, Steve Kehoe, Steve Satterfield, Sudarshan Govindaprasad, Sumit Gupta,
  Sungmin Cho, Sunny Virk, Suraj Subramanian, Sy~Choudhury, Sydney Goldman, Tal Remez, Tamar Glaser, Tamara Best, Thilo Kohler, Thomas Robinson, Tianhe Li, Tianjun Zhang, Tim Matthews, Timothy Chou, Tzook Shaked, Varun Vontimitta, Victoria Ajayi, Victoria Montanez, Vijai Mohan, Vinay~Satish Kumar, Vishal Mangla, Vítor Albiero, Vlad Ionescu, Vlad Poenaru, Vlad~Tiberiu Mihailescu, Vladimir Ivanov, Wei Li, Wenchen Wang, Wenwen Jiang, Wes Bouaziz, Will Constable, Xiaocheng Tang, Xiaofang Wang, Xiaojian Wu, Xiaolan Wang, Xide Xia, Xilun Wu, Xinbo Gao, Yanjun Chen, Ye~Hu, Ye~Jia, Ye~Qi, Yenda Li, Yilin Zhang, Ying Zhang, Yossi Adi, Youngjin Nam, Yu, Wang, Yuchen Hao, Yundi Qian, Yuzi He, Zach Rait, Zachary DeVito, Zef Rosnbrick, Zhaoduo Wen, Zhenyu Yang, and Zhiwei Zhao. 2024.
\newblock \href {https://arxiv.org/abs/2407.21783} {The llama 3 herd of models}.
\newblock \emph{Preprint}, arXiv:2407.21783.

\bibitem[{El~Mekki and Abdul-Mageed(2025)}]{self-mining}
Abdellah El~Mekki and Muhammad Abdul-Mageed. 2025.
\newblock \href {https://aclanthology.org/2025.findings-naacl.238/} {Effective self-mining of in-context examples for unsupervised machine translation with {LLM}s}.
\newblock In \emph{Findings of the Association for Computational Linguistics: NAACL 2025}, pages 4229--4256, Albuquerque, New Mexico. Association for Computational Linguistics.

\bibitem[{Goyal et~al.(2022)Goyal, Gao, Chaudhary, Chen, Wenzek, Ju, Krishnan, Ranzato, Guzm{\'a}n, and Fan}]{flores}
Naman Goyal, Cynthia Gao, Vishrav Chaudhary, Peng-Jen Chen, Guillaume Wenzek, Da~Ju, Sanjana Krishnan, Marc{'}Aurelio Ranzato, Francisco Guzm{\'a}n, and Angela Fan. 2022.
\newblock \href {https://doi.org/10.1162/tacl_a_00474} {The {F}lores-101 evaluation benchmark for low-resource and multilingual machine translation}.
\newblock \emph{Transactions of the Association for Computational Linguistics}, 10:522--538.

\bibitem[{Hao et~al.(2022)Hao, Sun, Dong, Han, Gu, and Wei}]{structured-prompting}
Yaru Hao, Yutao Sun, Li~Dong, Zhixiong Han, Yuxian Gu, and Furu Wei. 2022.
\newblock \href {https://doi.org/10.48550/ARXIV.2212.06713} {Structured prompting: Scaling in-context learning to 1, 000 examples}.
\newblock \emph{CoRR}, abs/2212.06713.

\bibitem[{Hendy et~al.(2023)Hendy, Abdelrehim, Sharaf, Raunak, Gabr, Matsushita, Kim, Afify, and Awadalla}]{how-good-hendy-2023}
Amr Hendy, Mohamed Abdelrehim, Amr Sharaf, Vikas Raunak, Mohamed Gabr, Hitokazu Matsushita, Young~Jin Kim, Mohamed Afify, and Hany~Hassan Awadalla. 2023.
\newblock \href {https://arxiv.org/abs/2302.09210} {How good are gpt models at machine translation? a comprehensive evaluation}.
\newblock \emph{Preprint}, arXiv:2302.09210.

\bibitem[{Ji et~al.(2024)Ji, Duan, Qiu, Zhang, Li, Yang, and Zhang}]{submodular}
Baijun Ji, Xiangyu Duan, Zhenyu Qiu, Tong Zhang, Junhui Li, Hao Yang, and Min Zhang. 2024.
\newblock \href {https://aclanthology.org/2024.lrec-main.1337} {Submodular-based in-context example selection for llms-based machine translation}.
\newblock In \emph{Proceedings of the 2024 Joint International Conference on Computational Linguistics, Language Resources and Evaluation, {LREC/COLING} 2024, 20-25 May, 2024, Torino, Italy}, pages 15398--15409. {ELRA} and {ICCL}.

\bibitem[{Jiao et~al.(2023)Jiao, Wang, Huang, Wang, and Tu}]{is-chatgpt-jiao-2023}
Wenxiang Jiao, Wenxuan Wang, Jen{-}tse Huang, Xing Wang, and Zhaopeng Tu. 2023.
\newblock \href {https://doi.org/10.48550/ARXIV.2301.08745} {Is chatgpt {A} good translator? {A} preliminary study}.
\newblock \emph{CoRR}, abs/2301.08745.

\bibitem[{Kim et~al.(2022)Kim, Cho, Kim, Kim, Yoo, and Lee}]{sg-icl}
Hyuhng~Joon Kim, Hyunsoo Cho, Junyeob Kim, Taeuk Kim, Kang~Min Yoo, and Sang{-}goo Lee. 2022.
\newblock \href {https://doi.org/10.48550/ARXIV.2206.08082} {Self-generated in-context learning: Leveraging auto-regressive language models as a demonstration generator}.
\newblock \emph{CoRR}, abs/2206.08082.

\bibitem[{Koneru et~al.(2024)Koneru, Exel, Huck, and Niehues}]{contextual-refinement-koneru-2024}
Sai Koneru, Miriam Exel, Matthias Huck, and Jan Niehues. 2024.
\newblock \href {https://doi.org/10.18653/V1/2024.NAACL-LONG.148} {Contextual refinement of translations: Large language models for sentence and document-level post-editing}.
\newblock In \emph{Proceedings of the 2024 Conference of the North American Chapter of the Association for Computational Linguistics: Human Language Technologies (Volume 1: Long Papers), {NAACL} 2024, Mexico City, Mexico, June 16-21, 2024}, pages 2711--2725. Association for Computational Linguistics.

\bibitem[{Kumar et~al.(2023)Kumar, Puduppully, Dabre, and Kunchukuttan}]{ctqscorer}
Aswanth Kumar, Ratish Puduppully, Raj Dabre, and Anoop Kunchukuttan. 2023.
\newblock \href {https://doi.org/10.18653/v1/2023.findings-emnlp.519} {{CTQS}corer: Combining multiple features for in-context example selection for machine translation}.
\newblock In \emph{Findings of the Association for Computational Linguistics: EMNLP 2023}, pages 7736--7752, Singapore. Association for Computational Linguistics.

\bibitem[{Li et~al.(2024)Li, Wang, and Li}]{human_gen_ICL}
Rui Li, Guoyin Wang, and Jiwei Li. 2024.
\newblock Are human-generated demonstrations necessary for in-context learning?
\newblock The Twelfth International Conference on Learning Representations.

\bibitem[{Lin et~al.(2022)Lin, Mihaylov, Artetxe, Wang, Chen, Simig, Ott, Goyal, Bhosale, Du, Pasunuru, Shleifer, Koura, Chaudhary, O{'}Horo, Wang, Zettlemoyer, Kozareva, Diab, Stoyanov, and Li}]{xglm}
Xi~Victoria Lin, Todor Mihaylov, Mikel Artetxe, Tianlu Wang, Shuohui Chen, Daniel Simig, Myle Ott, Naman Goyal, Shruti Bhosale, Jingfei Du, Ramakanth Pasunuru, Sam Shleifer, Punit~Singh Koura, Vishrav Chaudhary, Brian O{'}Horo, Jeff Wang, Luke Zettlemoyer, Zornitsa Kozareva, Mona Diab, Veselin Stoyanov, and Xian Li. 2022.
\newblock \href {https://doi.org/10.18653/v1/2022.emnlp-main.616} {Few-shot learning with multilingual generative language models}.
\newblock In \emph{Proceedings of the 2022 Conference on Empirical Methods in Natural Language Processing}, pages 9019--9052, Abu Dhabi, United Arab Emirates. Association for Computational Linguistics.

\bibitem[{Liu et~al.(2022)Liu, Shen, Zhang, Dolan, Carin, and Chen}]{icexample}
Jiachang Liu, Dinghan Shen, Yizhe Zhang, Bill Dolan, Lawrence Carin, and Weizhu Chen. 2022.
\newblock \href {https://doi.org/10.18653/v1/2022.deelio-1.10} {What makes good in-context examples for {GPT}-3?}
\newblock In \emph{Proceedings of Deep Learning Inside Out (DeeLIO 2022): The 3rd Workshop on Knowledge Extraction and Integration for Deep Learning Architectures}, pages 100--114, Dublin, Ireland and Online. Association for Computational Linguistics.

\bibitem[{Lu et~al.(2022)Lu, Bartolo, Moore, Riedel, and Stenetorp}]{overcoming}
Yao Lu, Max Bartolo, Alastair Moore, Sebastian Riedel, and Pontus Stenetorp. 2022.
\newblock \href {https://doi.org/10.18653/V1/2022.ACL-LONG.556} {Fantastically ordered prompts and where to find them: Overcoming few-shot prompt order sensitivity}.
\newblock In \emph{Proceedings of the 60th Annual Meeting of the Association for Computational Linguistics (Volume 1: Long Papers), {ACL} 2022, Dublin, Ireland, May 22-27, 2022}, pages 8086--8098. Association for Computational Linguistics.

\bibitem[{Lyu et~al.(2023)Lyu, Min, Beltagy, Zettlemoyer, and Hajishirzi}]{z-icl}
Xinxi Lyu, Sewon Min, Iz~Beltagy, Luke Zettlemoyer, and Hannaneh Hajishirzi. 2023.
\newblock \href {https://doi.org/10.18653/V1/2023.ACL-LONG.129} {{Z-ICL:} zero-shot in-context learning with pseudo-demonstrations}.
\newblock In \emph{Proceedings of the 61st Annual Meeting of the Association for Computational Linguistics (Volume 1: Long Papers), {ACL} 2023, Toronto, Canada, July 9-14, 2023}, pages 2304--2317. Association for Computational Linguistics.

\bibitem[{Min et~al.(2022)Min, Lyu, Holtzman, Artetxe, Lewis, Hajishirzi, and Zettlemoyer}]{rethinking}
Sewon Min, Xinxi Lyu, Ari Holtzman, Mikel Artetxe, Mike Lewis, Hannaneh Hajishirzi, and Luke Zettlemoyer. 2022.
\newblock \href {https://doi.org/10.18653/v1/2022.emnlp-main.759} {Rethinking the role of demonstrations: What makes in-context learning work?}
\newblock In \emph{Proceedings of the 2022 Conference on Empirical Methods in Natural Language Processing}, pages 11048--11064, Abu Dhabi, United Arab Emirates. Association for Computational Linguistics.

\bibitem[{Mosbach et~al.(2023)Mosbach, Pimentel, Ravfogel, Klakow, and Elazar}]{iclvsft}
Marius Mosbach, Tiago Pimentel, Shauli Ravfogel, Dietrich Klakow, and Yanai Elazar. 2023.
\newblock \href {https://doi.org/10.18653/V1/2023.FINDINGS-ACL.779} {Few-shot fine-tuning vs. in-context learning: {A} fair comparison and evaluation}.
\newblock In \emph{Findings of the Association for Computational Linguistics: {ACL} 2023, Toronto, Canada, July 9-14, 2023}, pages 12284--12314. Association for Computational Linguistics.

\bibitem[{Moslem et~al.(2022)Moslem, Haque, Kelleher, and Way}]{domain-specific-moslem-2022}
Yasmin Moslem, Rejwanul Haque, John Kelleher, and Andy Way. 2022.
\newblock \href {https://aclanthology.org/2022.amta-research.2} {Domain-specific text generation for machine translation}.
\newblock In \emph{Proceedings of the 15th biennial conference of the Association for Machine Translation in the Americas (Volume 1: Research Track)}, pages 14--30, Orlando, USA. Association for Machine Translation in the Americas.

\bibitem[{Moslem et~al.(2023)Moslem, Romani, Molaei, Kelleher, Haque, and Way}]{domain-terminology-moslem-2023}
Yasmin Moslem, Gianfranco Romani, Mahdi Molaei, John~D. Kelleher, Rejwanul Haque, and Andy Way. 2023.
\newblock \href {https://doi.org/10.18653/V1/2023.WMT-1.82} {Domain terminology integration into machine translation: Leveraging large language models}.
\newblock In \emph{Proceedings of the Eighth Conference on Machine Translation, {WMT} 2023, Singapore, December 6-7, 2023}, pages 902--911. Association for Computational Linguistics.

\bibitem[{Neubig et~al.(2019)Neubig, Dou, Hu, Michel, Pruthi, Wang, and Wieting}]{compare-mt}
Graham Neubig, Zi{-}Yi Dou, Junjie Hu, Paul Michel, Danish Pruthi, Xinyi Wang, and John Wieting. 2019.
\newblock \href {http://arxiv.org/abs/1903.07926} {compare-mt: {A} tool for holistic comparison of language generation systems}.
\newblock \emph{CoRR}, abs/1903.07926.

\bibitem[{{NLLB Team} et~al.(2022){NLLB Team}, Costa-jussà, Cross, Çelebi, Elbayad, Heafield, Heffernan, Kalbassi, Lam, Licht, Maillard, Sun, Wang, Wenzek, Youngblood, Akula, Barrault, Mejia-Gonzalez, Hansanti, Hoffman, Jarrett, Sadagopan, Rowe, Spruit, Tran, Andrews, Ayan, Bhosale, Edunov, Fan, Gao, Goswami, Guzmán, Koehn, Mourachko, Ropers, Saleem, Schwenk, and Wang}]{nllb2022}
{NLLB Team}, Marta~R. Costa-jussà, James Cross, Onur Çelebi, Maha Elbayad, Kenneth Heafield, Kevin Heffernan, Elahe Kalbassi, Janice Lam, Daniel Licht, Jean Maillard, Anna Sun, Skyler Wang, Guillaume Wenzek, Al~Youngblood, Bapi Akula, Loic Barrault, Gabriel Mejia-Gonzalez, Prangthip Hansanti, John Hoffman, Semarley Jarrett, Kaushik~Ram Sadagopan, Dirk Rowe, Shannon Spruit, Chau Tran, Pierre Andrews, Necip~Fazil Ayan, Shruti Bhosale, Sergey Edunov, Angela Fan, Cynthia Gao, Vedanuj Goswami, Francisco Guzmán, Philipp Koehn, Alexandre Mourachko, Christophe Ropers, Safiyyah Saleem, Holger Schwenk, and Jeff Wang. 2022.
\newblock No language left behind: Scaling human-centered machine translation.

\bibitem[{OpenAI(2023)}]{gpt4}
OpenAI. 2023.
\newblock \href {https://doi.org/10.48550/ARXIV.2303.08774} {{GPT-4} technical report}.
\newblock \emph{CoRR}, abs/2303.08774.

\bibitem[{Raunak et~al.(2023{\natexlab{a}})Raunak, Menezes, Post, and Hassan}]{literal}
Vikas Raunak, Arul Menezes, Matt Post, and Hany Hassan. 2023{\natexlab{a}}.
\newblock \href {https://doi.org/10.18653/V1/2023.ACL-SHORT.90} {Do gpts produce less literal translations?}
\newblock In \emph{Proceedings of the 61st Annual Meeting of the Association for Computational Linguistics (Volume 2: Short Papers), {ACL} 2023, Toronto, Canada, July 9-14, 2023}, pages 1041--1050. Association for Computational Linguistics.

\bibitem[{Raunak et~al.(2023{\natexlab{b}})Raunak, Sharaf, Wang, Awadalla, and Menezes}]{raunak-etal-2023-leveraging}
Vikas Raunak, Amr Sharaf, Yiren Wang, Hany Awadalla, and Arul Menezes. 2023{\natexlab{b}}.
\newblock \href {https://doi.org/10.18653/v1/2023.findings-emnlp.804} {Leveraging {GPT}-4 for automatic translation post-editing}.
\newblock In \emph{Findings of the Association for Computational Linguistics: EMNLP 2023}, pages 12009--12024, Singapore. Association for Computational Linguistics.

\bibitem[{Rei et~al.(2022{\natexlab{a}})Rei, C.~de Souza, Alves, Zerva, Farinha, Glushkova, Lavie, Coheur, and Martins}]{comet22}
Ricardo Rei, Jos{\'e}~G. C.~de Souza, Duarte Alves, Chrysoula Zerva, Ana~C Farinha, Taisiya Glushkova, Alon Lavie, Luisa Coheur, and Andr{\'e} F.~T. Martins. 2022{\natexlab{a}}.
\newblock \href {https://aclanthology.org/2022.wmt-1.52} {{COMET}-22: Unbabel-{IST} 2022 submission for the metrics shared task}.
\newblock In \emph{Proceedings of the Seventh Conference on Machine Translation (WMT)}, pages 578--585, Abu Dhabi, United Arab Emirates (Hybrid). Association for Computational Linguistics.

\bibitem[{Rei et~al.(2022{\natexlab{b}})Rei, Treviso, Guerreiro, Zerva, Farinha, Maroti, C.~de Souza, Glushkova, Alves, Coheur, Lavie, and Martins}]{rei-etal-2022-cometkiwi}
Ricardo Rei, Marcos Treviso, Nuno~M. Guerreiro, Chrysoula Zerva, Ana~C Farinha, Christine Maroti, Jos{\'e}~G. C.~de Souza, Taisiya Glushkova, Duarte Alves, Luisa Coheur, Alon Lavie, and Andr{\'e} F.~T. Martins. 2022{\natexlab{b}}.
\newblock \href {https://aclanthology.org/2022.wmt-1.60} {{C}omet{K}iwi: {IST}-unbabel 2022 submission for the quality estimation shared task}.
\newblock In \emph{Proceedings of the Seventh Conference on Machine Translation (WMT)}, pages 634--645, Abu Dhabi, United Arab Emirates (Hybrid). Association for Computational Linguistics.

\bibitem[{Rubin et~al.(2022)Rubin, Herzig, and Berant}]{retrieve-prompts}
Ohad Rubin, Jonathan Herzig, and Jonathan Berant. 2022.
\newblock \href {https://doi.org/10.18653/v1/2022.naacl-main.191} {Learning to retrieve prompts for in-context learning}.
\newblock In \emph{Proceedings of the 2022 Conference of the North American Chapter of the Association for Computational Linguistics: Human Language Technologies}, pages 2655--2671, Seattle, United States. Association for Computational Linguistics.

\bibitem[{Shin et~al.(2022)Shin, Lee, Ahn, Kim, Kim, Kim, Cho, Lee, Park, Ha, and Sung}]{ptcorpora}
Seongjin Shin, Sang{-}Woo Lee, Hwijeen Ahn, Sungdong Kim, HyoungSeok Kim, Boseop Kim, Kyunghyun Cho, Gichang Lee, Woo{-}Myoung Park, Jung{-}Woo Ha, and Nako Sung. 2022.
\newblock \href {https://doi.org/10.18653/V1/2022.NAACL-MAIN.380} {On the effect of pretraining corpora on in-context learning by a large-scale language model}.
\newblock In \emph{Proceedings of the 2022 Conference of the North American Chapter of the Association for Computational Linguistics: Human Language Technologies, {NAACL} 2022, Seattle, WA, United States, July 10-15, 2022}, pages 5168--5186. Association for Computational Linguistics.

\bibitem[{Sia and Duh(2023)}]{ontheflymt}
Suzanna Sia and Kevin Duh. 2023.
\newblock \href {https://aclanthology.org/2023.mtsummit-research.15} {In-context learning as maintaining coherency: A study of on-the-fly machine translation using large language models}.
\newblock In \emph{Proceedings of Machine Translation Summit XIX, Vol. 1: Research Track}, pages 173--185, Macau SAR, China. Asia-Pacific Association for Machine Translation.

\bibitem[{Su et~al.(2024)Su, Tai, Ji, Li, Yan, and Zhang}]{demonstration_augmentation}
Yi~Su, Yunpeng Tai, Yixin Ji, Juntao Li, Bowen Yan, and Min Zhang. 2024.
\newblock Demonstration augmentation for zero-shot in-context learning.
\newblock \emph{arXiv preprint arXiv:2406.01224}.

\bibitem[{Touvron et~al.(2023{\natexlab{a}})Touvron, Lavril, Izacard, Martinet, Lachaux, Lacroix, Rozi{\`{e}}re, Goyal, Hambro, Azhar, Rodriguez, Joulin, Grave, and Lample}]{llama}
Hugo Touvron, Thibaut Lavril, Gautier Izacard, Xavier Martinet, Marie{-}Anne Lachaux, Timoth{\'{e}}e Lacroix, Baptiste Rozi{\`{e}}re, Naman Goyal, Eric Hambro, Faisal Azhar, Aur{\'{e}}lien Rodriguez, Armand Joulin, Edouard Grave, and Guillaume Lample. 2023{\natexlab{a}}.
\newblock \href {https://doi.org/10.48550/ARXIV.2302.13971} {Llama: Open and efficient foundation language models}.
\newblock \emph{CoRR}, abs/2302.13971.

\bibitem[{Touvron et~al.(2023{\natexlab{b}})Touvron, Martin, Stone, Albert, Almahairi, Babaei, Bashlykov, Batra, Bhargava, Bhosale, Bikel, Blecher, Canton{-}Ferrer, Chen, Cucurull, Esiobu, Fernandes, Fu, Fu, Fuller, Gao, Goswami, Goyal, Hartshorn, Hosseini, Hou, Inan, Kardas, Kerkez, Khabsa, Kloumann, Korenev, Koura, Lachaux, Lavril, Lee, Liskovich, Lu, Mao, Martinet, Mihaylov, Mishra, Molybog, Nie, Poulton, Reizenstein, Rungta, Saladi, Schelten, Silva, Smith, Subramanian, Tan, Tang, Taylor, Williams, Kuan, Xu, Yan, Zarov, Zhang, Fan, Kambadur, Narang, Rodriguez, Stojnic, Edunov, and Scialom}]{llama2}
Hugo Touvron, Louis Martin, Kevin Stone, Peter Albert, Amjad Almahairi, Yasmine Babaei, Nikolay Bashlykov, Soumya Batra, Prajjwal Bhargava, Shruti Bhosale, Dan Bikel, Lukas Blecher, Cristian Canton{-}Ferrer, Moya Chen, Guillem Cucurull, David Esiobu, Jude Fernandes, Jeremy Fu, Wenyin Fu, Brian Fuller, Cynthia Gao, Vedanuj Goswami, Naman Goyal, Anthony Hartshorn, Saghar Hosseini, Rui Hou, Hakan Inan, Marcin Kardas, Viktor Kerkez, Madian Khabsa, Isabel Kloumann, Artem Korenev, Punit~Singh Koura, Marie{-}Anne Lachaux, Thibaut Lavril, Jenya Lee, Diana Liskovich, Yinghai Lu, Yuning Mao, Xavier Martinet, Todor Mihaylov, Pushkar Mishra, Igor Molybog, Yixin Nie, Andrew Poulton, Jeremy Reizenstein, Rashi Rungta, Kalyan Saladi, Alan Schelten, Ruan Silva, Eric~Michael Smith, Ranjan Subramanian, Xiaoqing~Ellen Tan, Binh Tang, Ross Taylor, Adina Williams, Jian~Xiang Kuan, Puxin Xu, Zheng Yan, Iliyan Zarov, Yuchen Zhang, Angela Fan, Melanie Kambadur, Sharan Narang, Aur{\'{e}}lien Rodriguez, Robert Stojnic, Sergey Edunov,
  and Thomas Scialom. 2023{\natexlab{b}}.
\newblock \href {https://doi.org/10.48550/ARXIV.2307.09288} {Llama 2: Open foundation and fine-tuned chat models}.
\newblock \emph{CoRR}, abs/2307.09288.

\bibitem[{Vilar et~al.(2023)Vilar, Freitag, Cherry, Luo, Ratnakar, and Foster}]{prompting-palm-vilar-2023}
David Vilar, Markus Freitag, Colin Cherry, Jiaming Luo, Viresh Ratnakar, and George Foster. 2023.
\newblock \href {https://doi.org/10.18653/v1/2023.acl-long.859} {Prompting {P}a{LM} for translation: Assessing strategies and performance}.
\newblock In \emph{Proceedings of the 61st Annual Meeting of the Association for Computational Linguistics (Volume 1: Long Papers)}, pages 15406--15427, Toronto, Canada. Association for Computational Linguistics.

\bibitem[{Wang et~al.(2024)Wang, Adelani, Agrawal, Masiak, Rei, Briakou, Carpuat, He, Bourhim, Bukula, Mohamed, Olatoye, Adewumi, Mokayed, Mwase, Kimotho, Yuehgoh, Aremu, Ojo, Muhammad, Osei, Omotayo, Chukwuneke, Ogayo, Hourrane, Anigri, Ndolela, Mangwana, Mohamed, Hassan, Awoyomi, Alkhaled, Al-Azzawi, Etori, Ochieng, Siro, Njoroge, Muchiri, Kimotho, Momo, Abolade, Ajao, Shode, Macharm, Iro, Abdullahi, Moore, Opoku, Akinjobi, Afolabi, Obiefuna, Ogbu, Brian, Otiende, Mbonu, Sari, Lu, and Stenetorp}]{africomet}
Jiayi Wang, David~Ifeoluwa Adelani, Sweta Agrawal, Marek Masiak, Ricardo Rei, Eleftheria Briakou, Marine Carpuat, Xuanli He, Sofia Bourhim, Andiswa Bukula, Muhidin Mohamed, Temitayo Olatoye, Tosin Adewumi, Hamam Mokayed, Christine Mwase, Wangui Kimotho, Foutse Yuehgoh, Anuoluwapo Aremu, Jessica Ojo, Shamsuddeen~Hassan Muhammad, Salomey Osei, Abdul-Hakeem Omotayo, Chiamaka Chukwuneke, Perez Ogayo, Oumaima Hourrane, Salma~El Anigri, Lolwethu Ndolela, Thabiso Mangwana, Shafie~Abdi Mohamed, Ayinde Hassan, Oluwabusayo~Olufunke Awoyomi, Lama Alkhaled, Sana Al-Azzawi, Naome~A. Etori, Millicent Ochieng, Clemencia Siro, Samuel Njoroge, Eric Muchiri, Wangari Kimotho, Lyse Naomi~Wamba Momo, Daud Abolade, Simbiat Ajao, Iyanuoluwa Shode, Ricky Macharm, Ruqayya~Nasir Iro, Saheed~S. Abdullahi, Stephen~E. Moore, Bernard Opoku, Zainab Akinjobi, Abeeb Afolabi, Nnaemeka Obiefuna, Onyekachi~Raphael Ogbu, Sam Brian, Verrah~Akinyi Otiende, Chinedu~Emmanuel Mbonu, Sakayo~Toadoum Sari, Yao Lu, and Pontus Stenetorp. 2024.
\newblock \href {https://arxiv.org/abs/2311.09828} {Afrimte and africomet: Enhancing comet to embrace under-resourced african languages}.
\newblock \emph{Preprint}, arXiv:2311.09828.

\bibitem[{Wei et~al.(2023)Wei, Wang, Schuurmans, Bosma, Ichter, Xia, Chi, Le, and Zhou}]{wei2023chainofthoughtpromptingelicitsreasoning}
Jason Wei, Xuezhi Wang, Dale Schuurmans, Maarten Bosma, Brian Ichter, Fei Xia, Ed~Chi, Quoc Le, and Denny Zhou. 2023.
\newblock \href {https://arxiv.org/abs/2201.11903} {Chain-of-thought prompting elicits reasoning in large language models}.
\newblock \emph{Preprint}, arXiv:2201.11903.

\bibitem[{Wolf et~al.(2020)Wolf, Debut, Sanh, Chaumond, Delangue, Moi, Cistac, Rault, Louf, Funtowicz, Davison, Shleifer, von Platen, Ma, Jernite, Plu, Xu, Le~Scao, Gugger, Drame, Lhoest, and Rush}]{huggingface}
Thomas Wolf, Lysandre Debut, Victor Sanh, Julien Chaumond, Clement Delangue, Anthony Moi, Pierric Cistac, Tim Rault, Remi Louf, Morgan Funtowicz, Joe Davison, Sam Shleifer, Patrick von Platen, Clara Ma, Yacine Jernite, Julien Plu, Canwen Xu, Teven Le~Scao, Sylvain Gugger, Mariama Drame, Quentin Lhoest, and Alexander Rush. 2020.
\newblock \href {https://doi.org/10.18653/v1/2020.emnlp-demos.6} {Transformers: State-of-the-art natural language processing}.
\newblock In \emph{Proceedings of the 2020 Conference on Empirical Methods in Natural Language Processing: System Demonstrations}, pages 38--45, Online. Association for Computational Linguistics.

\bibitem[{Xu et~al.(2024)Xu, Kim, Sharaf, and Awadalla}]{a-paradigm-xu-2024}
Haoran Xu, Young~Jin Kim, Amr Sharaf, and Hany~Hassan Awadalla. 2024.
\newblock \href {https://openreview.net/forum?id=farT6XXntP} {A paradigm shift in machine translation: Boosting translation performance of large language models}.
\newblock In \emph{The Twelfth International Conference on Learning Representations}.

\bibitem[{Yao et~al.(2022)Yao, Zhao, Yu, Du, Shafran, Narasimhan, and Cao}]{yao2022react}
Shunyu Yao, Jeffrey Zhao, Dian Yu, Nan Du, Izhak Shafran, Karthik Narasimhan, and Yuan Cao. 2022.
\newblock React: Synergizing reasoning and acting in language models.
\newblock \emph{arXiv preprint arXiv:2210.03629}.

\bibitem[{Zebaze et~al.(2025)Zebaze, Sagot, and Bawden}]{zebaze-etal-2025-context}
Armel~Randy Zebaze, Beno{\^i}t Sagot, and Rachel Bawden. 2025.
\newblock \href {https://aclanthology.org/2025.findings-naacl.68/} {In-context example selection via similarity search improves low-resource machine translation}.
\newblock In \emph{Findings of the Association for Computational Linguistics: NAACL 2025}, pages 1222--1252, Albuquerque, New Mexico. Association for Computational Linguistics.

\bibitem[{Zhang et~al.(2023)Zhang, Haddow, and Birch}]{promptingllm}
Biao Zhang, Barry Haddow, and Alexandra Birch. 2023.
\newblock \href {https://proceedings.mlr.press/v202/zhang23m.html} {Prompting large language model for machine translation: {A} case study}.
\newblock In \emph{International Conference on Machine Learning, {ICML} 2023, 23-29 July 2023, Honolulu, Hawaii, {USA}}, volume 202 of \emph{Proceedings of Machine Learning Research}, pages 41092--41110. {PMLR}.

\bibitem[{Zhang et~al.(2022)Zhang, Feng, and Tan}]{active}
Yiming Zhang, Shi Feng, and Chenhao Tan. 2022.
\newblock \href {https://doi.org/10.18653/v1/2022.emnlp-main.622} {Active example selection for in-context learning}.
\newblock In \emph{Proceedings of the 2022 Conference on Empirical Methods in Natural Language Processing}, pages 9134--9148, Abu Dhabi, United Arab Emirates. Association for Computational Linguistics.

\bibitem[{Zhao et~al.(2021)Zhao, Wallace, Feng, Klein, and Singh}]{calibrate}
Tony Zhao, Eric Wallace, Shi Feng, Dan Klein, and Sameer Singh. 2021.
\newblock \href {https://api.semanticscholar.org/CorpusID:231979430} {Calibrate before use: Improving few-shot performance of language models}.
\newblock In \emph{International Conference on Machine Learning}.

\end{thebibliography}

\clearpage
\appendix

\begin{figure*}[t!]
    \centering
    \includegraphics[width=0.9\textwidth]{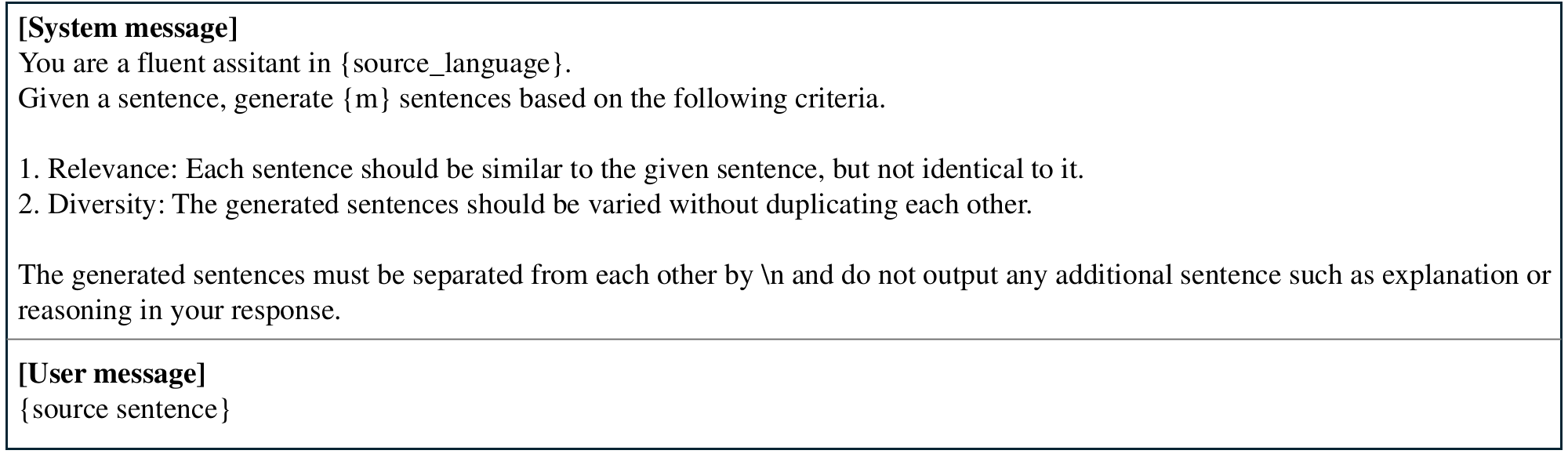}
    \caption{Prompt format used to generate source sentences with two criteria.}
    \label{fig:source_generation_prompt}
    
    \vspace{2em}
    
    \includegraphics[width=0.9\textwidth]{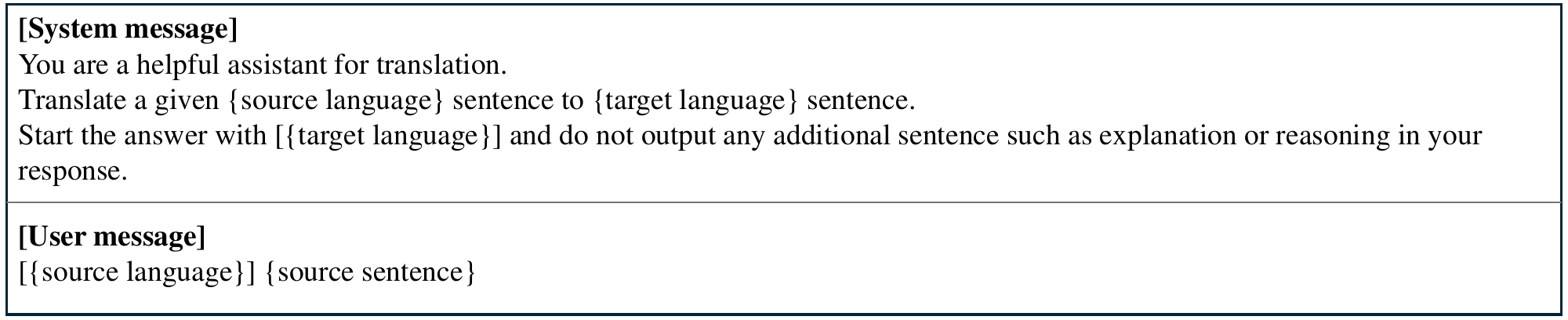}
    \caption{Prompt format used to translate a source sentence.}
    \label{fig:target_generation_prompt}
\end{figure*}

\section{Additional Information} \label{appendix:additional_information}
To ensure the reproducibility of the experiments, we used a temperature of 0.1 during token decoding and the dataset statistics can be found in Table~\ref{table:stat_flores}.
Experiments is mainly conducted with RTX 3090 GPU.
Zero-shot translation of the Flores devtest set using Llama-3.1-8B and 70B takes approximately 1 hour and 2 hours, respectively.
We load Llama-3.1-8B and 70B from Hugging Face's Transformers~\citep{huggingface}, with the LLaMA-3.1-70B model quantized to 4-bit.
The fixed pair set used in the experiment consists of the 1st, 2nd, 3rd, and 4th data points from the dev set.
We used ChatGPT to correct grammar.

\begin{table}[h]
\centering
\begin{tabular}{ccc}
\hline
train & dev & devtest \\ \hline
-     & 997 & 1012    \\ \hline
\end{tabular}
\caption{Data statistics of the Flores dataset~\citep{flores} used in all experiments.}
\label{table:stat_flores}
\end{table}
\end{document}